\documentclass[10pt,twocolumn,letterpaper]{article}

\usepackage{cvpr}              

\definecolor{cvprblue}{rgb}{0.21,0.49,0.74}
\usepackage[pagebackref,breaklinks,colorlinks,allcolors=cvprblue]{hyperref}

\def\paperID{3247} 
\def\confName{CVPR}
\def\confYear{2025}

\usepackage{graphicx}
\usepackage{xcolor}
\usepackage{colortbl}
\usepackage{multirow}
\definecolor{green1}{HTML}{b4dbca}
\definecolor{green2}{HTML}{daede4}
\definecolor{deepgreen}{HTML}{235d3a}
\definecolor{red1}{HTML}{ffa8a8}
\definecolor{red2}{HTML}{ffe3e3}
\definecolor{gray2}{HTML}{eeecea}

\definecolor{audio_red}{HTML}{D92523}
\definecolor{audio_purple}{HTML}{7262AC}
\definecolor{audio_blue}{HTML}{2E7EBB}

\title{EVOS: Efficient Implicit Neural Training via EVOlutionary Selector}

\author{
Weixiang Zhang\textsuperscript{\rm 1},
Shuzhao Xie\textsuperscript{\rm 1}, 
Chengwei Ren\textsuperscript{\rm 1},
Siyi Xie\textsuperscript{\rm 2},
Chen Tang\textsuperscript{\rm 3}, \\
Shijia Ge\textsuperscript{\rm 1}, 
Mingzi Wang\textsuperscript{\rm 1},
\vspace{0.2cm}
Zhi Wang\textsuperscript{\rm 1}\footnotemark[1] \\
\textsuperscript{\rm 1}Tsinghua Shenzhen International Graduate School, Tsinghua University \\
\textsuperscript{\rm 2}Xi'an Jiaotong University \:
\vspace{0.2cm}
\textsuperscript{\rm 3}The Chinese University of Hong Kong \\
\texttt{\href{https://weixiang-zhang.github.io/proj-evos/}{https://weixiang-zhang.github.io/proj-evos/}}
}

\begin{document}
\maketitle
{
\renewcommand{\thefootnote}{\fnsymbol{footnote}}
\footnotetext[1]{Corresponding author.}
}
\begin{abstract}
We propose EVOlutionary Selector (EVOS), an efficient training paradigm for accelerating Implicit Neural Representation (INR). Unlike conventional INR training that feeds all samples through the neural network in each iteration, our approach restricts training to strategically selected points, reducing computational overhead by eliminating redundant forward passes.
Specifically, we treat each sample as an individual in an evolutionary process, where only those fittest ones survive and merit inclusion in training, adaptively evolving with the neural network dynamics. While this is conceptually similar to Evolutionary Algorithms, their distinct objectives (selection for acceleration vs. iterative solution optimization) require a fundamental redefinition of evolutionary mechanisms for our context.
In response, we design sparse fitness evaluation, frequency-guided crossover, and augmented unbiased mutation to comprise EVOS. 
These components respectively guide sample selection with reduced computational cost, enhance performance through frequency-domain balance, and mitigate selection bias from cached evaluation. 
Extensive experiments demonstrate that our method achieves approximately 48\%-66\% reduction in training time while ensuring superior convergence without additional cost, establishing state-of-the-art acceleration among recent sampling-based strategies.
Our code is available at \href{https://github.com/zwx-open/EVOS-INR}{this link}.

\end{abstract}    
\section{Introduction}
Implicit Neural Representation~(INR) is an emerging paradigm~\cite{pe, siren, functa} that leverages Multilayer Perceptrons~(MLPs) to map spatial coordinates~(e.g., $(x,y)$ for images) to corresponding attributes~(e.g., RGB values of pixels), enabling continuous representation of diverse natural signals, including sketches~\cite{sketchinr}, audio~\cite{hypersound}, images~\cite{siren}, 3D shapes~\cite{deepsdf} and climate data~\cite{climate}. Compared to conventional discrete representations, INRs offer significant advantages in memory efficiency and inverse problem solving capability, demonstrating great potential in data compression~\cite{inr-compress}, image enhancement~\cite{inr-derain}, inverse imaging~\cite{inr-camera}, novel view synthesis~\cite{nerf}, and physics simulation~\cite{pinn}.

However, the encoding process of INRs remains computationally intensive due to dense iterations required for optimization, presenting a significant barrier to widespread adoption. Furthermore, spectral bias~\cite{bias}, where neural networks inherently prioritize low-frequency over high-frequency components, impedes fine detail fitting and introduces additional training overhead.
In response, various acceleration approaches have been proposed, each with inherent limitations: partition-based methods~\cite{acorn, miner, partition-acc} increase architectural complexity through multiple networks, explicit representation collaborations~\cite{diner, ingp, tensorf} compromise memory efficiency, and meta-learning approaches~\cite{meta-inr,meta-inr-trans} demand substantial homogeneous data for pre-training.

Unlike these approaches, recent sampling-based methods~\cite{egra,es,nmt} propose to accelerate training through sparsified forward passes by selectively sampling important coordinates~(samples). While this sampling paradigm offers potential cost-free acceleration, existing methods are limited by either relying solely on static signal structures without considering network dynamics~\cite{egra,nmt} or employing computationally expensive greedy algorithms for coordinate selection~\cite{nmt}, thus constraining their efficiency gains.

To address these limitations, we propose EVOlutionary Selector~(EVOS), a sampling-based method to accelerate INR training, inspired by principles of biological evolution.  
The basic motivation behind EVOS is that, the selection of optimal coordinate subsets conceptually parallels \textit{natural selection}, where only the fittest individuals survive and merit computational resources, adaptively evolving with network dynamics via \textit{crossover} and \textit{mutation} operations.
This paradigm shares principles with Evolutionary Algorithms (EA)~\cite{ec-ori}, yet serves a fundamentally different objective. While EA maintains a population of search points to iteratively refine candidate solutions toward optimal outcomes, our approach focuses on efficient subset selection rather than iterative solution optimization. Despite this distinction, evolutionary principles offer compelling advantages for complex~\cite{ec-complex} and dynamic optimization problems~\cite{ec-dynamic}, making them well-suited for addressing the non-convex nature of neural optimization~\cite{non-conv-mlp}. This motivates us to preserve the core evolutionary concept while redefining key components~(\textit{fitness evaluation}, \textit{crossover}, and \textit{mutation}) to align with INR acceleration goals.

Specifically, given a signal composed of coordinates $\mathbf{x}$ and corresponding attributes $\mathbf{y}$, the objective is to optimize an MLP $F_{\theta}(\mathbf{x})$ to approximate $\mathbf{y}$ with minimal loss. 
At each iteration $t$, EVOS treats a subset of coordinates $\mathbf{x^\prime}_t \subset \mathbf{x}$ as \textit{survivors}, which is selected through \textit{fitness evaluation} from the full coordinates $\mathbf{x}$. 
Subsequently, \textit{offspring} $\mathbf{z}_t$ are generated from these \textit{survivors} through \textit{crossover} and \textit{mutation} operations, and then fed into $F_{\theta}(\mathbf{z}_t)$ to achieve sparsified forward passes ($|\mathbf{z}_t| \leq |\mathbf{x}|$). 
The network parameters $\theta$ are then updated through backpropagation.
Through reducing the number of forward passes from $|\mathbf{x}|$ to $|\mathbf{z}_t|$ in each iteration, computational acceleration is achieved.

We redefine \textit{fitness evaluation}, \textit{crossover}, and \textit{mutation} as follows:
\textbf{(1)}~\underline{Sparse Fitness Evaluation}~(Sec.~\ref{fitness}). 
Computing fitness values in our context requires network forward passes for each coordinate, making conventional per-iteration fitness evaluation computationally prohibitive.
To address this, we perform sparse evaluations at strategic intervals and cache results for intermediate steps, significantly reducing computational overhead.
\textbf{(2)}~\underline{Frequency-Guided} \underline{Crossover}~(Sec.~\ref{crossover}). To address spectral bias that impedes high-frequency component fitting, we select coordinates from both low and high-frequency perspectives. These serve as parents for crossover operations, generating offspring that balance different frequency preferences and enhance performance with negligible computational cost.
\textbf{(3)}~\underline{Augmented Unbiased Mutation}~(Sec.~\ref{mutation}). Continuous reliance on cached evaluation results can introduce training bias and degrade performance~\cite{data-shuffle}. We mitigate this through stochastic incorporation of non-surviving points into the offspring set, introducing controlled uncertainty in each iteration through mutation-like operations.

We conduct extensive experiments to verify the performance~(Sec.\ref{sec:compare_sota}), compatibility~(Sec.~\ref{sec:backbones} \& Sec.~\ref{sec:network_size}) and effectiveness~(Sec.~\ref{sec:ablation_study}) of our method. 
With the integration of EVOS, the time cost of INRs' training can be reduced by 48\%-66\% without compromising reconstruction quality.
Compared to existing sampling-based methods, including recent approaches such as Soft Mining~\cite{soft} and INT~\cite{nmt}, EVOS achieves state-of-the-art training efficiency.
A key insight emerging from our study is that strategic sparsification of training samples not only reduces computational cost but also consistently enhances training performance, challenging the conventional paradigm of exhaustive data utilization for signal fitting tasks~\cite{siren,functa}.
To summarize, our contributions can be outlined as follows:
\begin{itemize}
    \item Inspired by principles of biological evolution, we propose EVOlutionary Selector~(EVOS), a sampling-based method that accelerates INR training while preserving reconstruction fidelity.
    \item We reformulate key notions in conventional evolutionary computation to meet our acceleration objectives, including sparse fitness evaluation, frequency-guided crossover, and augmented unbiased mutation.
    \item We conduct extensive experiments to verify the performance and compatibility of EVOS, demonstrating that EVOS achieves state-of-the-art efficiency improvement among recent sampling-based acceleration techniques.
\end{itemize}

\section{Related Work}
\begin{figure*}[!t]
    \centerline{\includegraphics[width=\textwidth]{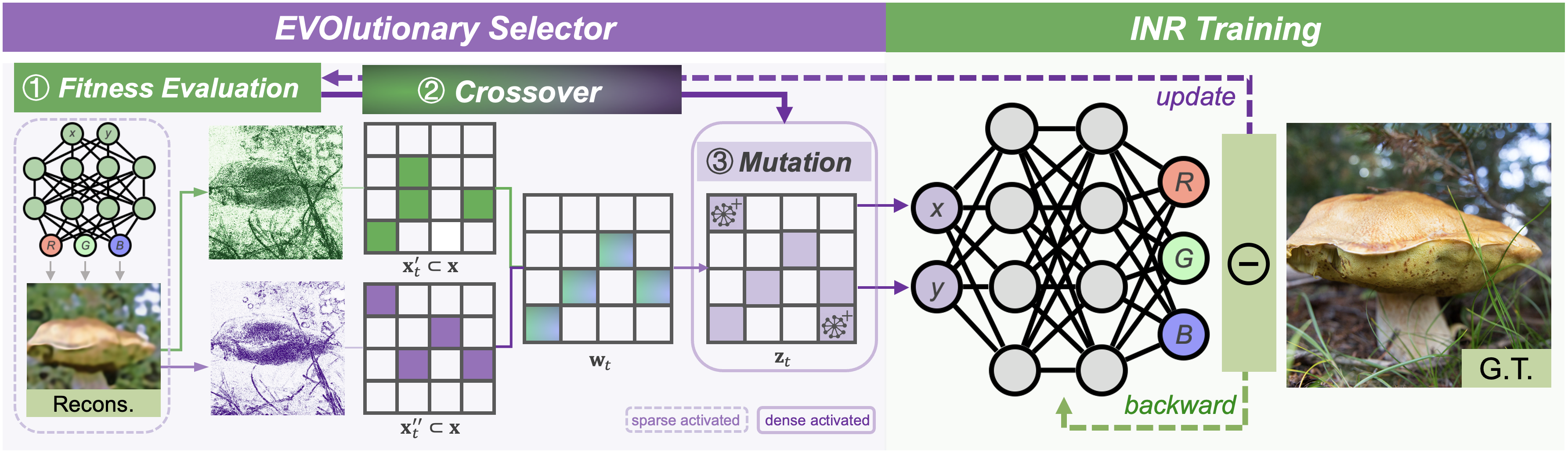}}
    \caption{\textbf{Overview of EVOS framework.} The proposed method aims to optimize a MLP for implicit signal representation via evolutionary selector. The process comprises three key components: (1) (Sparse) Fitness Evaluation~(Sec. \ref{fitness}) for efficiently guiding coordinate selection, (2) (Frequency-Guided) Crossover~(Sec. \ref{crossover}) for improving performance by balancing frequency domain preferences, and (3) (Augmented Unbiased) Mutation~(Sec. \ref{mutation}) for mitigating selection bias in each iteration. The selected coordinates from this evolutionary process are then fed into the network, enabling sparsified forward passes and reduced computational costs.}
    \label{fig:pipeline2}
\end{figure*}

\subsection{Implicit Neural Representations}
Implicit Neural Representations (INRs)~\cite{pe, siren, functa} continuously represent natural signals by training coordinate-based MLPs, whose structures are meticulously designed to achieve high representation capability, including position encoding~\cite{pe} and frequency-guided activation functions~\cite{siren, wire,gauss, finer}.
This representation paradigm has demonstrated broad applicability across diverse modalities, including images~\cite{siren, diner, coin}, videos~\cite{nerv, nvp, nirvana}, shapes~\cite{deepsdf, deeplocalshape, neuralangelo}, 3D scenes~\cite{nerf, ingp, tensorf}, and unconventional data formats~\cite{climate, fontinr, sketchinr}. Leveraging its memory efficiency and inherent capability in solving inverse problems, INRs have expanded beyond signal representation to applications in data compression~\cite{inr-compress}, generation~\cite{ddmi}, novel view synthesis~\cite{nerf}, inverse imaging~\cite{inr-camera}, and partial differential equation solving~\cite{pinn}.
Despite these advantages, the encoding process of INRs remains constrained by intensive computational requirements and spectral bias~\cite{bias}, which impedes high-fidelity detail reconstruction and limits their widespread adoption.

\subsection{Acceleration for INR Training}

Numerous approaches have been proposed to address the computational intensity of INR training, which can be categorized into four main streams:
\textbf{(1)} Partition-based methods decompose the input signal into smaller components, employing multiple compact MLPs for local representation. These include approaches based on regular blocks~\cite{kilonerf}, adaptive blocks~\cite{acorn}, Voronoi diagrams~\cite{derf}, Laplacian pyramids~\cite{miner}, and segmentation maps~\cite{partition-acc}.
\textbf{(2)} Prior-driven methods leverage knowledge distilled from large-scale datasets to expedite convergence. This category encompasses meta-learning approaches for parameter initialization~\cite{meta-inr, meta-inr-trans} and hyper-network-based weight generation~\cite{siren, hypersound}.
\textbf{(3)} Explicit caching methods trade spatial complexity for temporal efficiency by maintaining encoded features in explicit data structures, such as hash maps~\cite{ingp, diner}, point cloud~\cite{point-nerf}, low-rank tensors~\cite{tensorf}, and tree structures~\cite{tinc,nsvf}.
\textbf{(4)} Alternative approaches explore diverse strategies to enhance training efficiency, including modulator networks~\cite{modulate}, reparameterized training~\cite{reparam}, data transformation~\cite{search,sym-trans}, batch normalization~\cite{bn-inr}, and gradient adjustment~\cite{gradient-adjustment}.

Our method aligns with recent sampling-based acceleration approaches. Several methods have emerged in this direction: EGRA~\cite{egra}, Expansive Supervision~\cite{es}, and Soft Mining~\cite{soft}, tailored for Neural Radiance Field~\cite{nerf} (NeRF) training, demonstrate acceleration through selective \textit{Ray Sampling} based on edge detection, frequency priors, and Monte-Carlo methods, respectively. 
Most relevant to our approach, INT~\cite{nmt} addresses general INR acceleration by reformulating the learning process as a nonparametric teaching problem and employing greedy functional algorithms for coordinate selection.
EVOS distinctively leverages evolutionary principles for coordinate selection, introducing novel mechanisms for fitness evaluation, crossover, and mutation that achieve superior performance compared to existing sampling-based methods.

\section{Method}

\subsection{Formulation and Overview}
Consider a natural signal $S = (\mathbf{x}, \mathbf{y})$, where $\mathbf{x} \in \mathbb{R}^{m}$ and $\mathbf{y} \in \mathbb{R}^{n}$ represent coordinates and their corresponding attributes in $m$ and $n$ dimensions, respectively. Our target is to learn an INR by overfitting a MLP $F_\theta(\mathbf{x}): \mathbf{x} \in \mathbb{R}^m \mapsto \mathbf{y} \in \mathbb{R}^{n}$ to represent signal $S$.
Under the EVOS training paradigm, the optimization of $F_\theta$ is guided by a sparse forward pass $L(F_\theta(\textcolor{red}{\mathbf{z}_t}), \mathbf{y}_z)$ rather than the conventional $L(F_\theta(\textcolor{red}{\mathbf{x}}), \mathbf{y})$, where $\mathbf{z}_t \subset \mathbf{x}$ represents the selected coordinate subset at iteration $t$, and $\mathbf{y}_z \subset \mathbf{y}$ denotes their corresponding ground truth values. Here, $L(\cdot)$ represents the loss function.
The coordinate subset $\mathbf{z}_t$ is determined through three key steps:

\begin{enumerate}
    \item The \textit{survivors} $\mathbf{x^\prime}_t \subset \mathbf{x}$ are selected through Sparse Fitness Evaluation (Sec.\ref{fitness}), efficiently identifying the fittest candidates at a coarse level.
    \item The \textit{survivors} $\mathbf{x^\prime}_t$ undergo Frequency-Guided Crossover (Sec.\ref{crossover}) to generate \textit{offspring} $\mathbf{w}_t$, where parent selection is guided by frequency preferences to balance different spectral components.
    \item The \textit{offspring} $\mathbf{w}_t$ are further refined through Augmented Unbiased Mutation (Sec.\ref{mutation}) to obtain the final subset $\mathbf{z}_t$, incorporating stochastic perturbations to mitigate selection bias.
\end{enumerate}
Finally, the network parameters are optimized by gradients of $L(F_\theta(\mathbf{z}_t), \mathbf{y}_z)$. 
By reducing the number of forward passes from $|\mathbf{x}|$ to $|\mathbf{z}_t|$ in each iteration, EVOS achieves an acceleration factor of $I(\frac{|\mathbf{x}|-|\mathbf{z}_t|}{|\mathbf{x}|})$, where $I(\cdot)$ represents the ratio of forward pass time to total computation time. The overview of EVOS is illustrated in Fig.~\ref{fig:pipeline2}.

\subsection{Sparse Fitness Evaluation}
\label{fitness}

In conventional evolutionary algorithms, the fitness function evaluates the quality of each individual in the population, iteratively refining it until convergence to an optimal solution. Adapting this concept to EVOS, we define the fitness function as the distance between reconstructed and ground truth values, prioritizing coordinates with larger discrepancies for selection. Formally, the fitness function is expressed as $f(\mathbf{x}) = \mathcal{D}(F_\theta(\mathbf{x}), \mathbf{y})$, where $\mathcal{D}(\cdot)$ denotes the distance metric. The survivors $\mathbf{x^\prime}_t$ are selected through $\mathbf{x^\prime}_t = {\arg\max}_{{\mathbf{x}^{\prime}_t\subseteq \mathbf{x}}}(f(\mathbf{x}))$.

However, since implementing the fitness function~$f(\mathbf{x})$ requires computing all coordinates, performing such resource-intensive computations at each iteration would impose significant overhead, contradicting our efficiency objectives.
To address this, we introduce sparse fitness evaluation, which performs assessment only at key iterations while caching the results for intermediate steps. 
The sparse fitness function at iteration $t$ is formulated as:
\begin{equation}
\label{eq_sparse_fitness}
f_t(\mathbf{x}) = \Gamma(t)(\mathcal{D}(F_\theta(\mathbf{x}), \mathbf{y})) + (1- \Gamma(t))\hat{f}(\mathbf{x}),
\end{equation}
where $\hat{f}(\mathbf{x})$ represents the cached evaluation from the last key iteration and $\Gamma(t)$ is an indicator function that equals 1 at key iterations and 0 otherwise. When $\Gamma(t) = 1$, the cached evaluation is updated: $\hat{f}(\mathbf{x}){:=}f_t(\mathbf{x})$.

Notably, the frequency of key iterations in $\Gamma(t)$ presents a trade-off: more frequent evaluations offer finer-grained guidance but increase computational overhead. 
Our empirical analysis reveals that the distribution changes in $\mathcal{D}(F_\theta(\mathbf{x}), \mathbf{y})$ exhibit a gradual linear increase across iterations.
Based on this observation, we implement a linearly increasing schedule for key iterations to balance performance and efficiency:
\begin{equation}
\label{eq:eval_inter}
\Gamma(t) = 
\begin{cases}
1, & \ \text{if}~t~\bmod (\tau - \frac{\theta t}{T}) = 1  \\ 
0, & \ \text{otherwise}, 
\end{cases}
\end{equation}
where $T$ denotes the total number of iterations, and the interval between key iterations is $\tau -  \frac{\theta t}{T}$, controlled by coefficients $\tau$ and $\theta$. This interval decreases as training progresses, resulting in more frequent fitness evaluations.

\subsection{Frequency-Guided Crossover}

\label{crossover}

After fitness evaluation, the survivors $\mathbf{x}^\prime$ are selected using the fitness function $f(\mathbf{x}) = \mathcal{D}(F_\theta(\mathbf{x}), \mathbf{y})$. While defining $\mathcal{D}(\cdot)$ as the square error $||F_\theta(\mathbf{x})-\mathbf{y}||^2$ presents a straightforward measure of reconstruction quality, this coordinate-wise approach neglects spatial correlations, resulting in oversmoothed reconstructions and insufficient emphasis on high-frequency details. 
Given that INRs inherently exhibit spectral bias toward low-frequency components, this simple square error-based selection further exacerbates the challenge of learning high-frequency features.

To address this limitation, we propose a dual-perspective coordinate selection strategy that considers both low and high-frequency components. These two complementary subsets serve as parents in a crossover operation, producing offspring that balance different frequency preferences. Specifically, we employ the Laplacian operator $L_{lap}$, a differential operator based on the gradient divergence, to measure high-frequency components, while retaining square error for low-frequency assessment.
At iteration $t$, given coordinates $\mathbf{x}_t$, we compute two sets of parents: $\mathbf{x}^\prime_t$ (low-frequency preference) and $\mathbf{x}^{\prime\prime}_t$ (high-frequency preference) by selecting survivors using fitness functions $f^{low}_t(\mathbf{x})$ and $f^{high}_{t}(\mathbf{x})$, respectively:
\begin{equation}
\label{eq_parents}
\begin{aligned}
\mathbf{x}^\prime_t & = \underset{\{\mathbf{x}^\prime\}_k \subseteq \{\mathbf{x}\}_N} {\arg\max} ( \underbrace{||F_\theta(\mathbf{x})-\mathbf{y}||^2}_{f^{low}_t(\mathbf{x})}),  \\
\mathbf{x}^{\prime\prime}_t  & = \underset{\{\mathbf{x}^{\prime\prime}\}_k \subseteq \{\mathbf{x}\}_N} {\arg\max} ( \underbrace{||L_{lap}(F_\theta(\mathbf{x}))-L_{lap}(\mathbf{y})||^2}_{f^{high}_{t}(\mathbf{x})}),
\end{aligned}
\end{equation}
where $N$ denotes the total number of coordinates and $k$ represents the number of selected survivors. This selection process integrates with the sparse fitness evaluation mechanism described in Eq.~\ref{eq_sparse_fitness}.

Subsequently, the offspring $\mathbf{w}_t$ is generated through crossover operation between $\mathbf{x}^\prime_t$ and $\mathbf{x}^{\prime\prime}_t$:
\begin{equation}
\label{eq:cross-operation}
\mathbf{w}_t = \mathbf{x}^\prime_t \odot \mathbf{x}^{\prime\prime}_t = \{\mathbf{x}^\prime_t \cap \mathbf{x}^{\prime\prime}_t\} \cup \Psi(\mathbf{x}^\prime_t, \mathbf{x}^{\prime\prime}_t),
\end{equation}
where $\{\mathbf{x}^\prime_t \cap \mathbf{x}^{\prime\prime}_t\}$ represents coordinates of cross-frequency importance preserved in the offspring $\mathbf{w}_t$.
The balancer function $\Psi(\mathbf{x}^\prime_t, \mathbf{x}^{\prime\prime}_t)$ augments the offspring by selecting additional individuals from $\mathbf{x}^\prime_t$ and $\mathbf{x}^{\prime\prime}_t$ based on the current reconstruction quality across frequency components. This function is formulated as $\Psi(\mathbf{x}^\prime_t, \mathbf{x}^{\prime\prime}_t) = \mathbf{v}^\prime_t \cup \mathbf{v}^{\prime\prime}_t$, where:
\begin{equation}
\label{eq:v_sample}
\begin{aligned}
\mathbf{v}^\prime_t & \sim \mathcal{U}_{pl}(\mathbf{x}^\prime_t \setminus \mathbf{x}^{\prime\prime}_t), \\
\mathbf{v}^{\prime\prime}_t & \sim \mathcal{U}_{(1-p)l}(\mathbf{x}^{\prime\prime}_t \setminus \mathbf{x}^\prime_t).  
\end{aligned}
\end{equation}
Here, $p = \frac{f^{low}_t(\mathbf{x})}{(f^{low}_t(\mathbf{x}) + f^{high}_t(\mathbf{x}))}$, $l = k - |\mathbf{x}^\prime_t \cap \mathbf{x}^{\prime\prime}_t|$ denotes the number of samples, and $\mathcal{U}$ represents the uniform distribution. When low-frequency reconstruction exhibits superior quality~($p<0.5$), $\Psi(\mathbf{x}^\prime_t, \mathbf{x}^{\prime\prime}_t)$ prioritizes coordinates from the high-frequency preference set $\mathbf{x}^{\prime\prime}_t$, and vice versa.

{\textbf{Cross-Frequency Supervision.}}
To leverage the Laplacian gradient computed in Eq.~\ref{eq_parents}, we introduce cross-frequency loss to complement the crossover operation. For the mutated offspring $\mathbf{z}_t$ (Sec.~\ref{mutation}), we formulate the cross-frequency loss as:
\begin{equation}
\label{eq:cross-freq-loss}
\begin{aligned}
    L_t^{c} & = \lambda_l L_t^{l} + \lambda_h L_t^{h}, \\ 
    L_t^{l} & =   \ ||F_\theta(\mathbf{z}_t)-\mathbf{y}_z||^2, \\ 
    L_t^{h} & =   \ ||L_{lap}[F_\theta(\mathbf{z}_t) + \hat{F}_\theta(\mathbf{x} \setminus \mathbf{z}_t)]-L_{lap}(\mathbf{y})||^2.
\end{aligned}
\end{equation}
Since the $L_{lap}(\cdot)$ operator requires a full volume of attributes for its spatial convolution filter, the sparsified coordinates $\mathbf{z}_t$ cannot be directly applied. We address this by incorporating cached evaluation results $\hat{F}_\theta(\mathbf{x} \setminus \mathbf{z}_t)$ from the last key iteration.
$\lambda_l$ and $\lambda_h$ denote the scaling factors.

\begin{figure}[!t]
    \centerline{\includegraphics[width=\columnwidth]{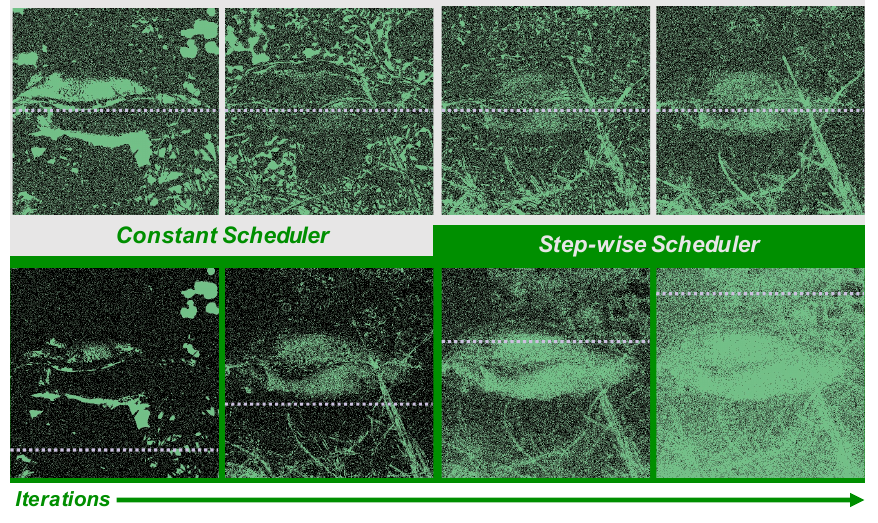}}
    \caption{Visualization of Constant (Top) and Step-wise (Bottom) schedulers. Constant maintains fixed selection ratio throughout iterations, while Step-wise implements progressive increase.}
    \label{fig:epoch}
\end{figure}

\subsection{Augmented Unbiased Mutation}
\label{mutation}
While EVOS's sparse fitness evaluation (Sec.~\ref{fitness}) significantly reduce computational cost, they introduce a potential drawback: the network receives deterministic offspring coordinates $\mathbf{w}_t$ during consecutive iterations, potentially inducing selection bias and degrading performance~\cite{data-shuffle}. To address this, we incorporate uncertainty through mutation, stochastically integrating a small portion of non-selected individuals $\{\mathbf{x} \setminus \mathbf{w}_t\}$ into the offspring. This strategy maintains population diversity and mitigates training bias.

At iteration $t$, given offspring set $\mathbf{w}_t$, we generate mutated coordinates $\mathbf{z}_t$ through the augmented unbiased mutation operator $\mathcal{M}(\cdot)$:
\begin{equation}
\label{eq_mutation}
\mathbf{z}_t = \mathcal{M}(\mathbf{w}_t) = \mathbf{w}_t \cup \mathbf{m}_t, \ \mathbf{m}_t \sim \mathcal{U}_{\alpha k}(\{\mathbf{x} \setminus \mathbf{w}_t \}),
\end{equation}
where $\mathbf{m}_t$ denotes mutated coordinates of length $\alpha k$ uniformly sampled from the non-selected set $\{\mathbf{x} \setminus \mathbf{w}_t\}$, and $\alpha$ represents the mutation ratio controlling uncertainty. 
The final number of mutated offspring is $|\mathbf{z}_t| = (1+\alpha)k < N$.
Given the negligible computational cost of uniform sampling, Eq.~\ref{eq_mutation} is executed in each iteration.

Finally, we utilize the mutated offspring $\mathbf{z}_t$ as the fittest subset of coordinates for iteration $t$ in training network $F_\theta$. The network parameters $\theta$ are optimized according to:
\begin{equation}
\theta \mathrel{:=} \theta - \eta \nabla L_t^c(F_\theta(\mathbf{z}_t), \mathbf{y}_z),
\end{equation}
where $\eta$ denotes the learning rate and $L_t^c$ represents the \textit{cross-frequency loss} detailed in Eq.~\ref{eq:cross-freq-loss}.

\subsection{Discussions}

\textbf{Scheduler for Selection Size.} 
For determining the selection intensity $q$ ($q=|\mathbf{z}_t|$), we implement two scheduling mechanisms:
\textbf{(1)} \underline{Constant scheduler}: maintains a fixed selection size $q=\beta N$ throughout training, where $\beta \in (0,1)$ and $N$ denotes the total number of coordinates $\mathbf{x}$, ensuring consistent memory efficiency.
\textbf{(2)} \underline{Step-wise scheduler}: progressively increases $q$ following a step function during training, which achieves optimal performance as demonstrated in INT~\cite{nmt}. We adopt this scheduler with default parameters with INT ($q$ increasing from $0.2N$ to $N$).
Both schedulers are implemented by adjusting $k$ (Eq.~\ref{eq_parents}) according to $q=(1+\alpha)k$.
Fig.~\ref{fig:epoch} illustrates the temporal dynamics of both scheduling mechanisms.

\noindent \textbf{Complexity.} Our method achieves acceleration without introducing additional storage overhead. Selection operations in EVOS account for merely 1.34\% of training time, which is negligible considering the method's 48\%-66\% reduction in total training duration.

\section{Experiments}
\label{sec:exp}

\begin{table*}[tbp]
    
    \centering
    \begin{tabular}{l|ccc|ccc|ccc|c}
        \toprule
                & \multicolumn{3}{c|}{1k Iterations} & \multicolumn{3}{c|}{2k Iterations} & \multicolumn{3}{c|}{5k Iterations} & \multicolumn{1}{c}{Time$\downarrow$} \\ 
        Strategies & PSNR $\uparrow$ &SSIM $\uparrow$ &LPIPS$\downarrow$ &  PSNR$\uparrow$ & SSIM$\uparrow$& LPIPS$\downarrow$ & PSNR$\uparrow$& SSIM$\uparrow$& LPIPS$\downarrow$& (sec) \\
        \midrule
        \rowcolor{gray2}
        Standard   & 31.06 & 0.899 & 0.123 & 34.34 & 0.944 & 0.042 & 37.10 & 0.964 & 0.021 & 180.45 \\
        \midrule
        Uniform.  & 30.35 & 0.885 & 0.149 & 33.33 & 0.932 & 0.058 & 36.11 & 0.956 & 0.026 & \cellcolor{green2}94.13 \\
        
        EGRA~\cite{egra}  & 30.37 & 0.885 & 0.150 & 33.40 &0.932 & 0.057 &  36.22 & 0.957 & 0.025 & \cellcolor{green2}103.31  \\

        Expan.~\cite{es}  & 30.95 & 0.893 & \cellcolor{green2}0.121 & 33.70 & 0.933 & 0.049 & 36.28 &  0.955 & 0.023 & \cellcolor{green2}96.39 \\ 

        Soft Mining~\cite{soft}  & \cellcolor{green2}31.49 & \cellcolor{green2}0.903 & 0.124 & 33.52 & 0.931 & 0.072 & 35.31 & 0.948 & 0.045 & \cellcolor{green2}100.35 \\
    
        INT~\cite{nmt}~(incre.) & \cellcolor{green2}31.67  & \cellcolor{green2}0.901 & \cellcolor{green2}0.117 & 31.95 & 0.894  & 0.096 & 34.62 & 0.923 & 0.059 & \cellcolor{green2}110.55 \\
        
        
        INT~\cite{nmt}~(dense.) & \cellcolor{green2}31.67 & \cellcolor{green2}0.901 & \cellcolor{green2}0.117 & \cellcolor{green2}34.53 & 0.938 & 0.052 &  \cellcolor{green2}37.21 & 0.959 &  0.027 & 161.89\\
        
        EVOS~(w/o CFS.) &  \cellcolor{green2}31.64 & 0.899 & \cellcolor{green2}0.107 & \cellcolor{green2}34.73 & 0.940 & \cellcolor{green2}0.040 & \cellcolor{green2}37.49 & 0.960 & \cellcolor{green2}0.019 & \cellcolor{green2}95.43 \\ 
        
        EVOS~(proposed) & \cellcolor{green1}32.71 & \cellcolor{green1}0.912 & \cellcolor{green1}0.082 & \cellcolor{green1}35.89 & \cellcolor{green1}0.951 & \cellcolor{green1}0.031 & \cellcolor{green2}37.81 & 0.962 & \cellcolor{green2}0.018 & \cellcolor{green2}97.39 \\ 
        
        \midrule
        \underline{EGRA}~\cite{egra}  & 29.59 &  0.867 & 0.187  & 33.01 & 0.927 &  0.068 & 37.08 & 0.963 & \cellcolor{green2}0.020 & \cellcolor{green2}105.71  \\ 
       
        \underline{Expansive Sup.}~\cite{es}  & 30.58 & 0.883 & 0.138 & 33.55 & 0.930 & 0.058 & 36.95 & 0.960 & 0.026 & \cellcolor{green2}103.09\\ 
        \underline{INT}~\cite{nmt}~(incre.)\textsuperscript{$\dagger$} & \cellcolor{green2}31.52 & 0.880 & 0.137 & 30.59 & 0.863 & 0.138 & \cellcolor{green2}37.17 & 0.962 & \cellcolor{green2}0.020 &  \cellcolor{green2}124.35 \\

        \underline{EVOS}~(w/o CFS.) & 30.99 &   0.873  &   0.124   &  \cellcolor{green2}34.49  &   0.930  &  0.050   &  \cellcolor{green2}37.81  & \cellcolor{green2}0.965 &  \cellcolor{green2}0.019 & \cellcolor{green2}105.82 \\
        
        \underline{EVOS}~(proposed) & \cellcolor{green2}31.69 & 0.882 & \cellcolor{green2}0.106 &  \cellcolor{green2}35.61 &    0.940  &   \cellcolor{green2}0.041  &  \cellcolor{green1}38.43  &   \cellcolor{green1}0.968  &  \cellcolor{green1}0.016 &  \cellcolor{green2}108.60 \\ 
        
        \bottomrule
        \multicolumn{8}{l}{\scriptsize{$\dagger$ denotes the best-performing variant reported in INT~\cite{nmt}.}}
         
    \end{tabular}
    \caption
    {{Comparison of sampling strategies under fixed iterations.}
    Strategies without underlines employ constant scheduler ($\beta=0.5$), while
    \underline{underlined strategies} implement step-wise scheduler. \colorbox{green1}{Forest}: the best performance; \colorbox{green2}{Mint}: exceeds standard training.}
    \label{table:compare_0.5}
\end{table*}

\begin{table}[tbp]

    \small
    \centering
    \begin{tabular}{l|>{\columncolor{gray2}}c|cccc|>{\columncolor{green2}}c}
        \toprule
        PSNR & Stand. & EGRA & Expan. & Soft. & INT\textsuperscript{$\dagger$}  & EVOS  \\ 
        \midrule
        25 dB & 9.04 & 6.19 & 5.08& 4.99 & 6.15 & 4.73\\
        30 dB & 30.75 & 19.63 & 16.42& 14.14 & 18.68 & 12.48\\ 
        35 dB & 88.64 & 67.15 & 59.41& 84.81 & 75.35  & 29.73 \\ 
        \bottomrule
        
    \end{tabular}
    \caption{
    {{Comparison of computational time~(seconds) required to achieve target PSNR across different strategies.}} 
    }
    \label{table:target_psnr}
\end{table}

\subsection{Implement Details}

\noindent\textbf{Experimental Settings.} 
We set the total number of iterations $T$ to 5000, with hyperparameters $\tau=100$, $\theta=0.01$, $\alpha=0.5$, $\lambda_l=1$, $\lambda_h=1\text{e-}5$, and $\beta=0.5$.
Following \cite{pe,finer}, we evaluated our method on processed DIV2K datasets~\cite{div2k}
using 3$\times$256 MLP with SIREN~\cite{siren} architecture unless otherwise stated.
Additional experiments across various modalities and datasets are provided in the {\textit{supplementary materials}}.
All experiments in this section were performed using the PyTorch framework~\cite{torch} on 4 NVIDIA RTX 3090 GPUs with 24.58 GB VRAM each.

\noindent\textbf{Evaluation Mechanisms.} 
We evaluated reconstruction quality using PSNR, SSIM~\cite{ssim}, and LPIPS~\cite{lpips} metrics. 
The efficiency improvements were assessed through two mechanisms: 
\textbf{(1)} comparing time required to achieve target reconstruction quality, and 
\textbf{(2)} comparing performance and computational cost under fixed iterations. 
The first directly reveals efficiency gains across different strategies, while the second compares the per-iteration performance between full data utilization and partial data sampling.

\subsection{Comparison with State-of-the-arts Strategies}
\label{sec:compare_sota}

\noindent \textbf{Settings.}
We compared our method against Uniform Sampling~(Uniform.), EGRA~\cite{egra}, Expansive Supervision~(Expan.)~\cite{es}, Soft Mining~\cite{soft}, INT~\cite{nmt}, and conventional full-coordinate training (Standard) commonly used in INR encoding~\cite{pe,siren,wire,gauss,finer}. 
For INT, which addresses the identical application scenario as ours, we strictly followed to its official implementation and settings.
Although EGRA and Expan. were originally tailored for NeRF training, their straightforward mechanisms enabled direct adaptation to our experiments.
For Soft Mining, we modified parameters to accommodate general INR training, though it demonstrated limited effectiveness in 2D image fitting tasks.
Through empirical tuning, optimal performance was achieved with $\alpha_{soft}=0.5$ and disabled fuzzy indexing (detailed in supplementary materials).
Apart from these necessary adjustments, all experimental settings remained consistent across methods.

\noindent \textbf{Quantitative Results.}
Tables~\ref{table:compare_0.5} and~\ref{table:target_psnr} present quantitative comparisons under fixed iterations and target PSNR benchmarks, respectively. 
In Table~\ref{table:compare_0.5}, INT~(incre.) and INT~(dense.) denote variants optimized for minimal time cost and maximal reconstruction quality, respectively. 
For fair comparison independent of loss function variations, we include EVOS (w/o CFS.), which implements vanilla L2 loss without Cross-Frequency Supervision (Eq.~\ref{eq:cross-freq-loss}).
Our method demonstrates consistent superior efficiency across all comparative strategies. 
Specifically, under fixed iterations, our approach achieves 39.82\% time reduction while improving PSNR by 1.33 dB compared to standard training (final row of Table~\ref{table:compare_0.5}).
Furthermore, when targeting PSNR values of 25, 30, and 35 dB, our method reduces computational time by \underline{47.68\%}, \underline{59.41\%}, and \underline{66.46\%} respectively (Table~\ref{table:target_psnr}), outperforming all alternative approaches.

\begin{figure}[!t]
    \centerline{\includegraphics[width=\columnwidth]{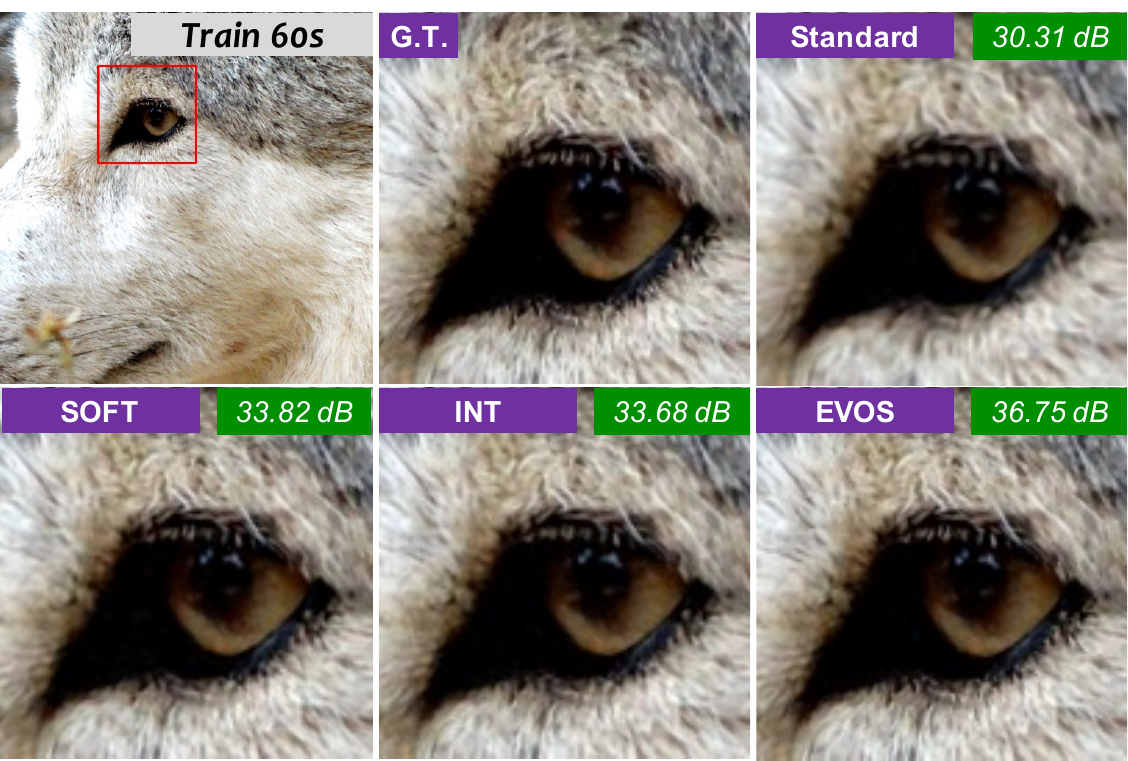}}
    \caption{Visual comparison of sampling-based acceleration methods for INR training. All methods are evaluated under identical conditions with a fixed training duration of 60 seconds.}
    \label{fig:visual_compare}
\end{figure}

\noindent \textbf{Visualization Results.}
Fig.~\ref{fig:visual_compare} presents reconstructed images after 60 seconds of training using different acceleration methods. Under identical time constraints, our method achieves superior reconstruction quality, surpassing alternative approaches by 2.31 to 6.44 dB. Detailed examination reveals that our method particularly excels in preserving high-fidelity details, as evidenced by the wolf's eye and fur regions, producing results closest to ground truth.

\begin{figure}[!t]
    \centerline{\includegraphics[width=\columnwidth]{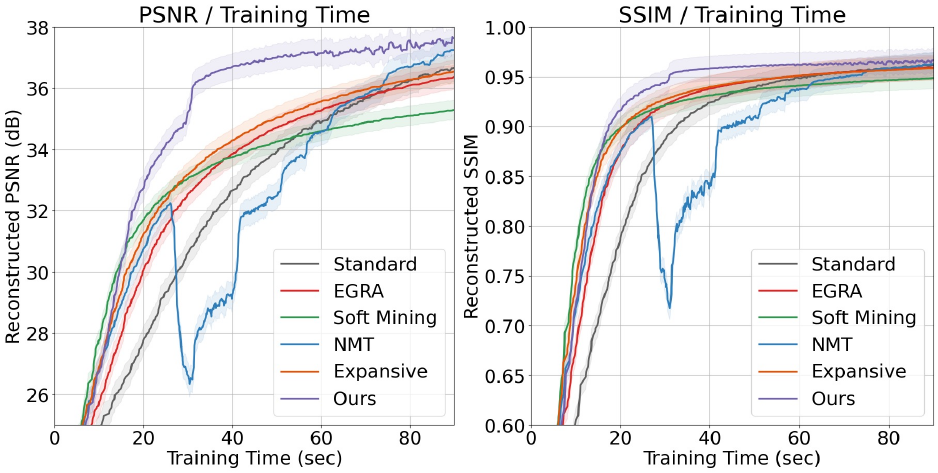}}
    \caption{Comparison of reconstruction quality (PSNR \& SSIM) across different sampling-based methods over training time.}
    \label{fig:loss}
\end{figure}

\noindent \textbf{Temporal Evolution of Training.}
The temporal evolution of reconstruction quality is illustrated in Fig.~\ref{fig:loss}, which plots PSNR and SSIM metrics against training time. For precise evaluation, additional full-coordinate forward passes were performed after each optimization step to compute accurate metrics, with these evaluation passes excluded from the total training time. 
Our method demonstrates both faster convergence and superior performance compared to alternative approaches, particularly in PSNR metrics. 
Moreover, EVOS exhibits more stable performance gains than INT, maintaining its advantage throughout the training process.

\noindent \textbf{Analysis of EGRA and Expan.}
While EGRA and Expan. leverage edge detection and frequency priors for coordinate sampling, accelerating training in the early stages, their overreliance on static image information without considering dynamic network changes results in suboptimal long-term performance.

\noindent \textbf{Analysis of Soft Mining.}
Despite empirical parameter tuning, Soft Mining exhibits deteriorating reconstruction quality, particularly in later training stages. This limitation likely stems from sampling space incompatibility. While Soft Mining effectively samples approximately 5\% of coordinates from vast 3D radiance field data in NeRF training, conventional INR training typically processes all available data in each iteration due to relatively small data volume. This fundamental difference in sampling requirements explains its reduced effectiveness in non-NeRF INR applications. Supporting experimental evidence is provided in the supplementary material.

\noindent \textbf{Analysis of INT.}
INT, which closely aligns with our application context, employs a greedy algorithm to select high-priority coordinates.
While its best-performing variant \mbox{\underline{INT}(incre.)\textsuperscript{$\dagger$}} demonstrates improvements over standard training (31\% time savings, 0.07 dB PSNR gain), it exhibits unstable reconstruction quality, as shown in Fig.~\ref{fig:loss}. The performance degradation in early stages stems from abrupt reductions in MLP evaluation frequency inherent to its greedy algorithm. EVOS circumvents this limitation through our sparse fitness evaluation design. Moreover, our crossover and mutation mechanisms achieve superior performance compared to INT's greedy selection approach.

\noindent \textbf{Sample Weighting ALSO Benefits ``Overfitting''.}
An intuitive assumption suggests that sparsifying training samples would compromise per-iteration reconstruction quality while reducing computational cost, due to incomplete data utilization. 
However, the results of EVOS (w/o CFS) in Table~\ref{table:compare_0.5} challenge this intuition, demonstrating that our strategy achieves superior fitting quality compared to full-data training under identical iteration counts.
This phenomenon can be understood through the lens of sample weighting~\cite{stablenet,marw}, a technique that improves model generality by adjusting sample observation frequencies during training.
Indeed, EVOS can be viewed as a specialized form of sample weighting that reweights signal coordinates during signal fitting, implicitly regularizing the loss function through selective sampling.
Notably, this finding extends the benefits of sample weighting beyond traditional model generalization to signal fitting tasks (inherently an overfitting scenario without test set validation), revealing an intriguing contradiction that merits further investigation.

\begin{figure*}[!t]
    \centerline{\includegraphics[width=\textwidth]{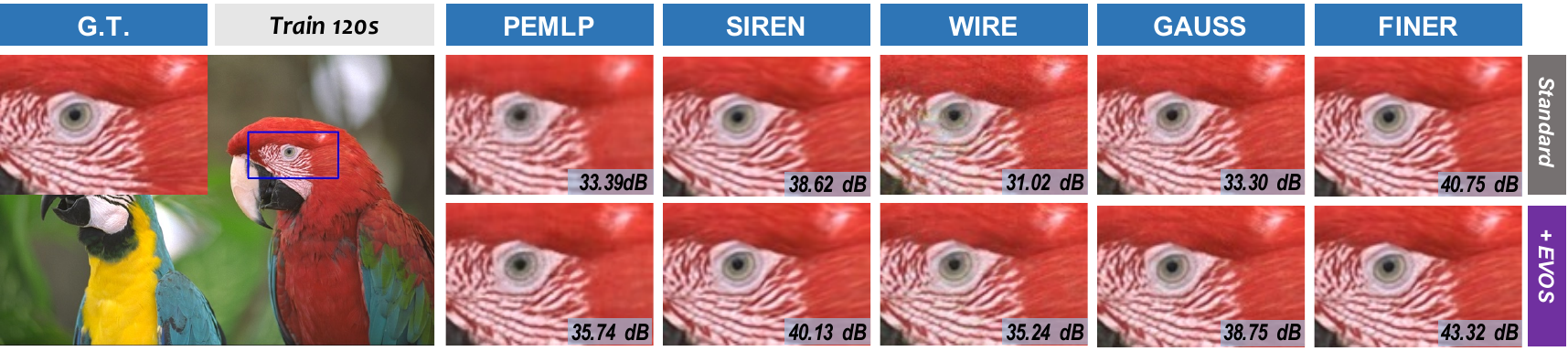}}
    \caption{Visualization of performance improvements across different backbone architectures when integrated EVOS. All experiments are conducted under consistent conditions with a fixed training duration of 120 seconds.}
    \label{fig:backbone_compare}
\end{figure*}

\subsection{Compatibility for Different Backbones}
\label{sec:backbones}
To evaluate EVOS's compatibility, we conducted experiments across multiple INR architectures: PEMLP~\cite{pe}, WIRE~\cite{wire}, GAUSS~\cite{gauss}, and FINER~\cite{finer}. These architectures represent diverse approaches to enhancing INR expressiveness through innovations in frequency encoding and activation functions.
Following~\cite{finer}, we implemented a consistent learning rate scheduler across all backbones while maintaining their official initial learning rates. 
The quantitative results in Table~\ref{table:backbone-exp} demonstrate EVOS's consistent acceleration capabilities across different architectures, showing substantial efficiency gains and performance improvements. In most cases (highlighted in \colorbox{green2}{green}), EVOS achieves simultaneous improvements in both efficiency and reconstruction quality. 
Fig.~\ref{fig:backbone_compare} illustrates the reconstruction quality across different backbones after 120 seconds of training. EVOS integration consistently enhances detail preservation across all architectures, particularly in the fine features of the parrot's eye region.

\begin{table}[tbp]
    
   
    \small
    \centering
    \begin{tabular}{l|ccc|c|c}
        \toprule
          &  \multicolumn{3}{c|}{PSNR} & SSIM & Time \\ 
          Backbones & (1k) $\uparrow$ & (2k) $\uparrow$ & (5k) $\uparrow$ & (5k) $\uparrow$ & (sec) $\downarrow$ \\ 
        \midrule
        PEMLP~\cite{pe} & 25.99 & 27.61 & 29.43 & 0.841 & 148.84 \\
        PEMLP+EVOS & \cellcolor{green2}26.10 & \cellcolor{white}28.00 & \cellcolor{green2}30.04 & \cellcolor{green2}0.842 & \cellcolor{green2}83.18 \\ 
        \midrule
        SIREN~\cite{siren} & 30.37 & 33.35 & 36.24 & 0.960 & 177.26 \\
        SIREN+EVOS & \cellcolor{green2}32.12 & \cellcolor{green2}35.10 & \cellcolor{green2}36.84 & \cellcolor{green2}0.961 & \cellcolor{green2}100.43 \\
        \midrule
        WIRE~\cite{wire} & 28.96 & 31.17 & 33.72 & 0.918 & 553.31\\
        WIRE+EVOS & 28.59 & \cellcolor{green2}32.91 & \cellcolor{green2}34.56 & \cellcolor{green2}0.927 & \cellcolor{green2}291.30 \\
        \midrule
        GAUSS~\cite{gauss} & 30.38 & 32.83 & 35.48 & 0.945 & 244.91 \\
        GAUSS+EVOS & 30.21 & \cellcolor{green2}34.36 & \cellcolor{green2}35.48 & 0.941 & \cellcolor{green2}133.28 \\
        \midrule
        FINER~\cite{finer} & 36.93 & 38.67 & 40.72 & 0.978 & 260.81 \\
        FINER+EVOS & 36.49 & \cellcolor{green2}39.54 & 40.61 &  0.976 & \cellcolor{green2}123.60 \\

        \bottomrule
    \end{tabular}
    \caption{
    {Quantitative result across different backbones.} 
    }
    \label{table:backbone-exp}
\end{table}

\subsection{Compatibility for Different Network Sizes}
\label{sec:network_size}
To evaluate EVOS's compatibility with varying network architectures, we tested its performance across MLP configurations ranging from 1$\times$64 to 3$\times$256. Table~\ref{table:network_size_exp} presents primary results, demonstrating EVOS's effectiveness in improving training efficiency across diverse architectures, with efficiency gains becoming more pronounced as network size increases. From these observations, we draw two key conclusions:
\textbf{(1)} EVOS demonstrates strong generalization across network architectures, enabling efficient training acceleration for varying INR expressiveness requirements.
\textbf{(2)} The effectiveness of EVOS increases with network size, suggesting that larger architectures exhibit greater redundancy which our method effectively addresses.

\begin{table}[tbp]
    \small
    \centering
    \begin{tabular}{l|cccc|c}
        \toprule
          & \multicolumn{4}{c|}{PSNR}  & Time \\ 
        Sizes & (1k) $\uparrow$ & (2k) $\uparrow$ & (5k) $\uparrow$ & (10k) $\uparrow$ & (sec) $\downarrow$ \\ 
        \midrule
        1$\times$64 & 20.60 & 21.49 & 22.31 & 22.58 & 47.71\\
        1$\times$64~({EVOS}) & \cellcolor{green2}20.65 & \cellcolor{green2}21.53 & 22.26 & 22.54 & \cellcolor{green2}37.28 \\ 
        \midrule
        2$\times$64 & 22.57 & 23.71 & 24.57 &  24.79 & 72.19 \\
        2$\times$64~({EVOS}) & \cellcolor{green2}22.72 & \cellcolor{green2}23.85 & 24.56 & 24.77 & \cellcolor{green2}47.18  \\ 
        \midrule
        2$\times$128 & 25.29 & 26.97 & 28.33 &  28.82 & 114.04  \\
        \rowcolor{green2}
        \cellcolor{white}2$\times$128~({EVOS}) & 25.68 & 27.29 & 28.40 & 28.83 &  74.59 \\ 
        \midrule
        2$\times$256 & 28.25 &  30.30 & 32.27 & 33.30 & 262.46 \\
        \rowcolor{green2}
        \cellcolor{white}2$\times$256~({EVOS}) & 29.05 &  31.09 & 32.51 & 33.49 & 144.98 \\ 
        \midrule
        3$\times$256 & 31.06 & 34.34 & 37.10 & 38.60 & 366.30 \\
        \rowcolor{green2}
        \cellcolor{white}3$\times$256~({EVOS}) & 32.71 & 36.09 & 37.81 & 38.71   & 254.33\\ 
        \bottomrule
        
    \end{tabular}

     \caption{
    {{Quantitative result across different network sizes.}} 
    }
    \label{table:network_size_exp}
\end{table}

\subsection{Ablation Study}
\label{sec:ablation_study}
\noindent \textbf{Settings.}
We conducted comprehensive ablation studies to evaluate the quantitative contributions of EVOS components: Sparse Fitness Evaluation (Eval.), Frequency-Guided Crossover (Cross.), and Augmented Unbiased Mutation (Mutat.). 
Beyond component-level analysis, we examined key design parameters including the interval parameter $\tau$ and linear increasing coefficient $\theta$ in Eq.~\ref{eq:eval_inter}, Cross-Frequency Supervision (CFS, Eq.~\ref{eq:cross-freq-loss}), and mutation ratio $\alpha$ in Eq.~\ref{eq_mutation}. All experiments followed the settings in Sec.~\ref{sec:compare_sota}, varying only the parameter under investigation.

\noindent \textbf{Results and Analysis.}
Table~\ref{table:ablation} presents the result of ablation experiment.
The absence of Eval.~(Sec. \ref{fitness}) results in unguided selection, significantly degrading early-stage performance. Without Mutat.~(Sec. \ref{mutation}), selection bias from cached selections substantially impairs final performance. Components from Cross.~(Sec. \ref{crossover}), CFS~{Eq. \ref{eq:cross-freq-loss}}, and the linear increasing indicator~(Eq. \ref{eq:eval_inter}) enhance performance with minimal computational overhead.
Interestingly, increasing evaluation frequency (reducing $\tau$ to 1 or 10) does not improve performance despite higher computational costs. We suspect that overly frequent evaluation induces excessive training instability, making it difficult for INRs to consistently capture signal features during the training process.

\begin{table}[tbp]

    \small
    \centering
    \begin{tabular}{l|cc|c|c}
        \toprule
          & \multicolumn{2}{c|}{PSNR} & SSIM & Time \\ 
         Settings & (1k) $\uparrow$ & (5k) $\uparrow$ & (5k) $\uparrow$ & (sec) $\downarrow$ \\ 
        
        \midrule
        \rowcolor{green2}
        EVOS & 32.71  & 37.81 &  0.962 & 97.39  \\
        \midrule
        w/o Eval.~(Sec. \ref{fitness}) & \cellcolor{red1}20.80  & 35.41 & 0.952 & 93.41  \\
        $\theta = 0$~(Eq.\ref{eq:eval_inter}) & 32.69  & 37.54 & 0.961 & 95.26 \\ 
        $\tau= 1$ 
        ~(Eq. \ref{eq:eval_inter}) & 31.57  & 35.51 & 0.943 & \cellcolor{red1}142.39 \\ 
        $\tau= 10$ ~(Eq. \ref{eq:eval_inter}) & 32.42  & 36.37 & 0.953 & 105.23 \\ 
        
        \midrule
        w/o Cross.~(Sec. \ref{crossover}) & 31.67  & 37.04 & 0.956 &  94.35 \\ 
        w/o CFS.~(Eq. \ref{eq:cross-freq-loss}) & 31.64 &  37.49 & 0.960 & 95.43 \\ 
        \midrule
        w/o Mutat.~(Sec. \ref{mutation}) & 31.18  & \cellcolor{red1}34.94 & \cellcolor{red1}0.921 & 97.13  \\
        $\alpha = 0.1$~(Eq. \ref{eq_mutation}) & 32.33  & 37.11 & 0.953 & 96.64 \\
         $\alpha = 1.0$~(Eq. \ref{eq_mutation}) & 32.10  & 37.14 & 0.958&101.51 \\ 
        \bottomrule
    \end{tabular}
   \caption{
    {{Ablation study.} \colorbox{red1}{Coral}: the worst performance. }
    }
    \label{table:ablation}
\end{table}
\section{Conclusion}
\label{sec:conclusion}
In this paper, we propose EVOlutionary Selector (EVOS) to accelerate INR training. 
Extensive experiments demonstrate EVOS's superior performance, surpassing existing sampling-based acceleration methods. Notably, we discover that strategic sparsification of training samples not only reduces computational cost but also improves per-iteration performance, suggesting that sample weighting techniques can enhance optimization in signal fitting tasks beyond their traditional role in model generalization.

\clearpage
\section*{Acknowledgments}
This work is supported in part by National Key Research and Development
Project of China~(Grant No. 2023YFF0905502), National Natural Science Foundation of China~(Grant No. 92467204 and 62472249), and Shenzhen Science and Technology Program~(Grant No. JCYJ20220818101014030 and KJZD20240903102300001).
We thank anonymous reviewers for their valuable advice.

{
    \small
    \bibliographystyle{ieeenat_fullname}
    \bibliography{main}
}

\clearpage
\clearpage

\setcounter{page}{1}
\setcounter{section}{0}
\onecolumn
\renewcommand{\thesection}{\Alph{section}}

\maketitlesupplementary

\section{1D Audio Fitting Task}
\textbf{Background \& Settings.} 
Fitting 1D audio data can be formulated as $F_\theta(\mathbf{x}): (t) \mapsto (a)$, where $a$ represents the amplitude value at time step $t$.
For this task, we utilized the \textit{test.clean} split from LibriSpeech~\cite{libri} dataset, with each audio sample truncated to the initial 5 seconds at a 16,000 Hz sampling rate.
Following Siamese SIREN~\cite{siamese}, we configured both $\omega$ and $\omega_0$ to 100 in the SIREN architecture. 
The quality of reconstructed audio was evaluated using SI-SNR, STOI~\cite{STOI}, PESQ~\cite{pesq}, and Mean Square Error~(MSE) metrics. 
Due to task incompatibility issues exhibited by EGRA~\cite{egra}, Expan.~\cite{es}, and Soft Mining~\cite{soft}, we restricted our experiments to standard training, uniform sampling, INT~\cite{nmt}, and its variants.

\begin{figure}[t]
    \centerline{\includegraphics[width=\columnwidth]{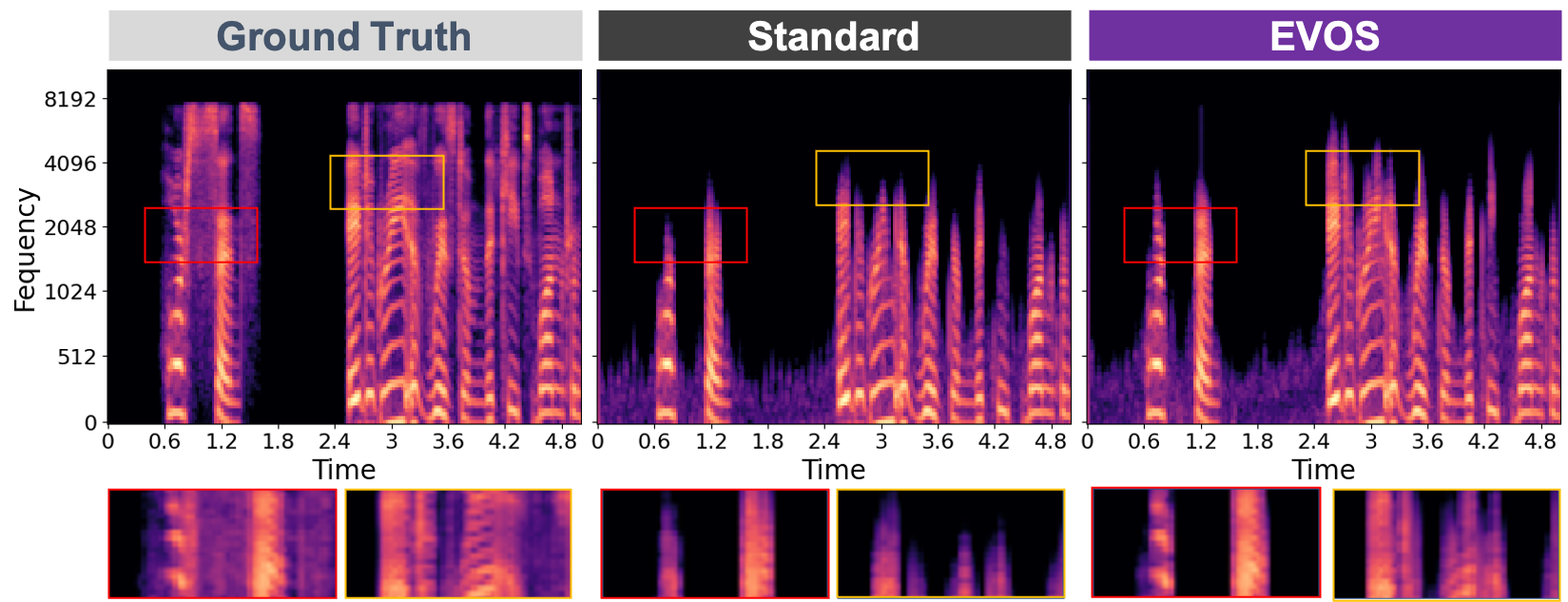}}
    \caption{Visual comparison of Mel spectrogram reconstructions with 30-second training duration. Due to spectral bias, INRs exhibit lower expressiveness in high-frequency regions compared to low-frequency regions. EVOS integration can alleviate this limitation under fixed time constraints.}
    \label{fig:visual_mel}
\end{figure}

\begin{figure}[t]
    \centerline{\includegraphics[width=\columnwidth]{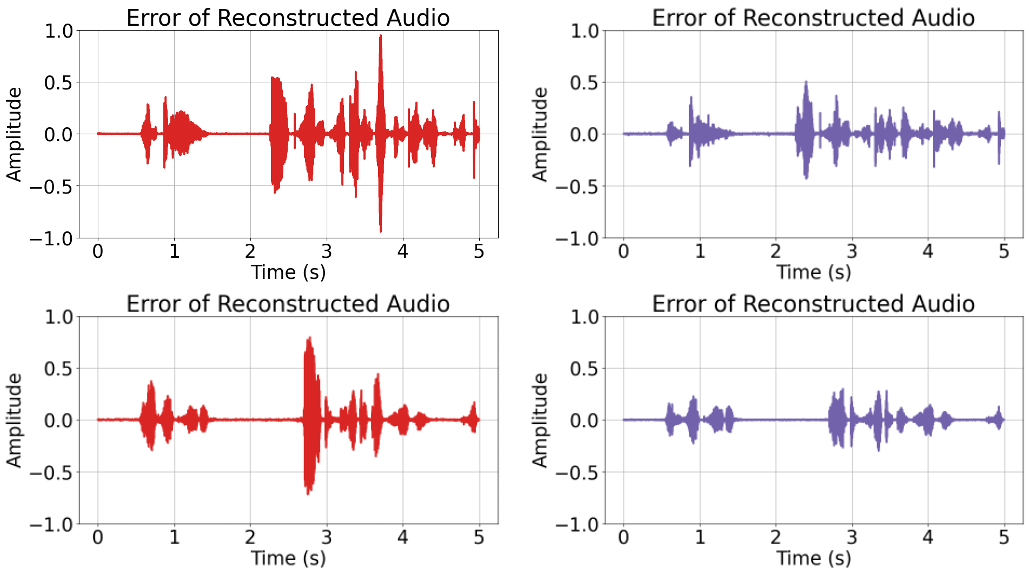}}
    \caption{Visual comparison of reconstruction error for 1D audio fitting with 30-second training duration. \textbf{\textcolor{audio_red}{Red}} and \textbf{\textcolor{audio_purple}{Purple}} lines represent standard training and our method, respectively.}
    \label{fig:audio_error}
\end{figure}

\noindent \textbf{Results.} 
As shown in Table~\ref{table:audio_sample}, our method simultaneously achieves reduced training time and enhanced reconstruction quality per iteration compared to INT and its variants, consistently outperforming standard training across all metrics. 
Fig.~\ref{fig:audio_error} illustrates the reconstruction error under a fixed 30-second training duration, comparing standard training (red) with our method (purple). Our approach demonstrates notably lower error rates than standard training. 
The Mel spectrogram visualization in Fig.~\ref{fig:visual_mel} further validates the effectiveness of our method.

\section{2D Text Fitting Task}
\textbf{Background \& Settings.} 
Fitting 2D synthesized text image data can be formulated as $F_\theta(\mathbf{x}): (x,y) \mapsto (r,g,b)$. 
The dataset in this experiment was obtained from~\cite{pe}. Such synthesized text data differs from natural images in its inherently imbalanced distribution and limited pixel intensity variety.
For example, a synthesized text image containing three words might only have four intensity values (three text colors and one background color), unlike natural images with abundant color variations.
Given the task's relative simplicity, we set the total iterations $T=1000$ and used SIREN as the backbone, while maintaining other settings consistent with Sec.~\ref{sec:compare_sota}.

\begin{figure}[t]
    \centerline{\includegraphics[width=\columnwidth]{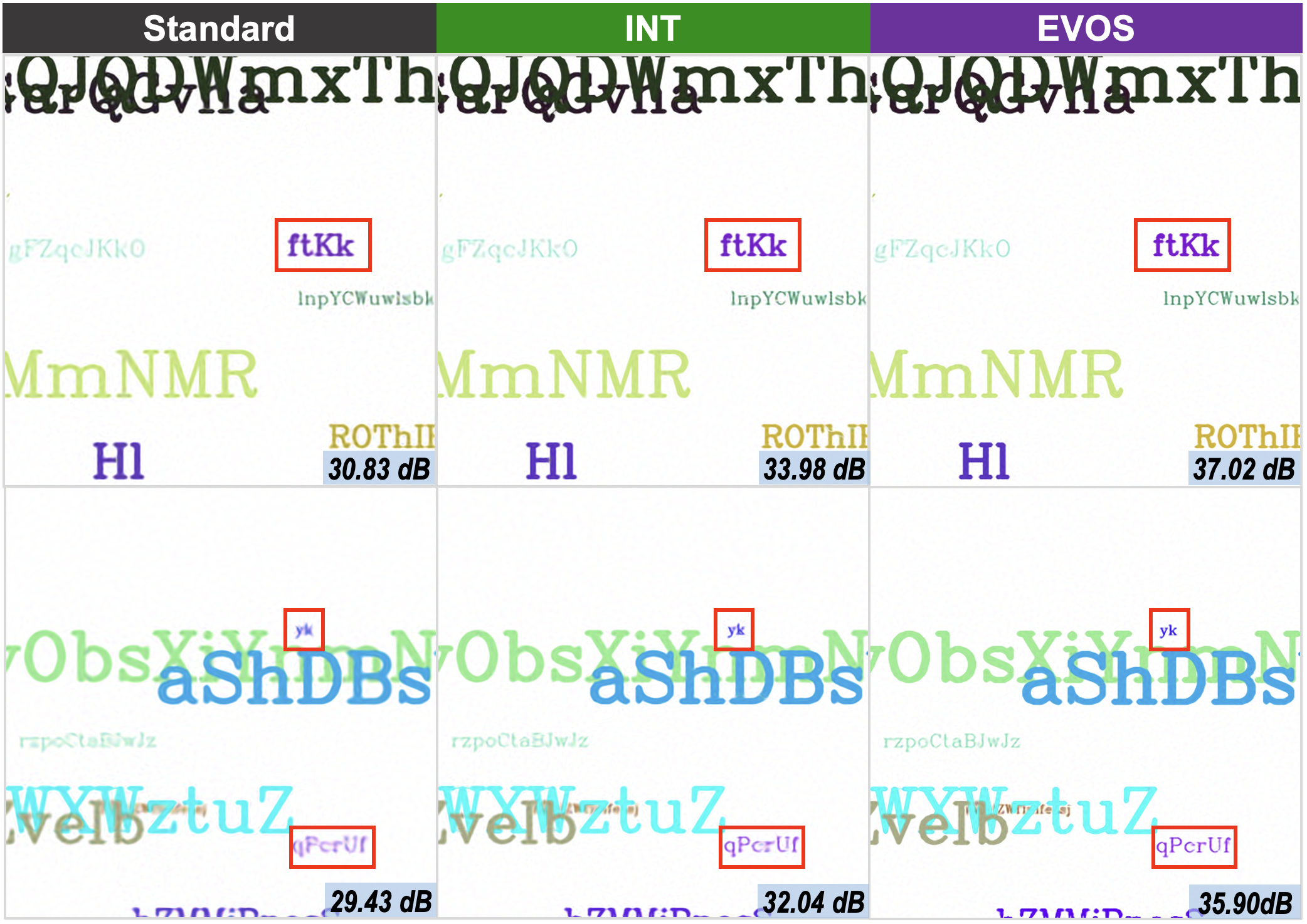}}
    \caption{Visual comparison for 2D text fitting task. We employ INT~(dense.) rather than INT\textsuperscript{$\dagger$} due to the latter's performance degradation in this task. All experiments are conducted under consistent conditions with a fixed training duration of 20 seconds.}
    \label{fig:text_compare}
\end{figure}

\noindent \textbf{Results.} 
Table~\ref{table:text_sample} demonstrates that our method achieves the most significant efficiency improvements compared to existing sampling methods~\cite{egra,es,nmt,soft} on 2D text fitting tasks. Table~\ref{table:text_backbone} verifies our method's compatibility across different architectures~\cite{siren,gauss,pe,finer}. Furthermore, Fig.~\ref{fig:text_compare} presents reconstructed texts after 20 seconds of training using various acceleration methods. Since INT\textsuperscript{$\dagger$} shows degraded performance in this task, we used INT~(dense) for comparison. The highlighted regions (red boxes) demonstrate our method's superior reconstruction quality.

\begin{table}[tbp]
    \small
    \centering
    \begin{tabular}{l|cccc|c}
        \toprule
            & SI-SNR & STOI &  PESQ &  MSE& Time \\ 
        Strategies   & $\uparrow$ & $\uparrow$ & $\downarrow$ & $\downarrow$~(e-4) & $\downarrow$ (sec) \\ 
        \midrule
        Standard  & 11.21 & 0.896 & 1.387 & 2.835& 47.62\\ 
        \midrule
        Uniform. & 9.88 & 0.855 &  1.274 & 4.938 & 28.47 \\
        \midrule
        INT~\cite{nmt} & 11.62 & 0.904 & 1.307 & 2.410 & 31.36\\ 
        INT~\cite{nmt}\textsuperscript{$\star$} & 12.14 & 0.908 & 1.409 & 2.067 & 45.80 \\
        \underline{INT}~\cite{nmt}\textsuperscript{$\dagger$} & 11.79 & 0.905 & 1.394 & 2.233 & 33.42\\ 
        \midrule
        EVOS  & 12.35 & 0.910 & 1.411 & 2.014 & 29.44\\
        \underline{EVOS} &  \cellcolor{green1}12.95 & \cellcolor{green1}0.921 & \cellcolor{green1}1.449 & \cellcolor{green1}1.660 & 31.63 \\ 
        \bottomrule
        \multicolumn{6}{l}{\scriptsize{$\star$ denotes INT~(dense.)}} \\ 
        \multicolumn{6}{l}{\scriptsize{$\dagger$ denotes the best-performing variant reported in INT~\cite{nmt}.}} \\ 
    \end{tabular}
    \caption{
    {Comparison of sampling strategies on 1D audio fitting.}
    \colorbox{green1}{Forest}: the best performance.
    }
    \label{table:audio_sample}
\end{table}

\begin{table}[tbp]
    
   
    \small
    \centering
    \begin{tabular}{l|cccc|c}
        \toprule
           & PSNR & SSIM & LPIPS &  MSE  & Time \\ 
        Strategies  & $\uparrow$ & $\uparrow$ & $\downarrow$ & $\downarrow$~(e-3) & $\downarrow$ (sec) \\ 
    
        \midrule
        Standard  & 35.15 & 0.986 & 0.022 & 1.527 & 35.89 \\ 
        \midrule
        Uniform. & 33.73 & 0.983 &  0.031 & 2.118 & 20.68\\
        EGRA~\cite{egra} & 34.07 & 0.983 & 0.029 & 1.979 & 21.01 \\ 
        Expan.~\cite{es} &  36.92 & 0.984 & 0.017 &  1.043 & 20.33\\
        Soft.~\cite{soft} & 37.01 & 0.981 & 0.144 & 1.053 & 22.18\\
        \midrule
        INT~\cite{nmt}\textsuperscript{$\star$}  & 36.76 & \cellcolor{green1}0.989 &  0.155 & 1.088 & 31.58\\
        \underline{INT}~\cite{nmt}\textsuperscript{$\dagger$} & 35.59 & 0.987 & 0.020 & 1.389 & 24.75\\ 
        \midrule
        EVOS  & \cellcolor{green1}37.42 & 0.985 & \cellcolor{green1}0.016 & \cellcolor{green1}1.002 & 24.15 \\

        \bottomrule
        \multicolumn{6}{l}{\scriptsize{$\star$ denotes INT~(dense.)}} \\ 
        \multicolumn{6}{l}{\scriptsize{$\dagger$ denotes the best-performing variant reported in INT~\cite{nmt}.}} \\ 
    \end{tabular}
    \caption{
    {Comparison of sampling strategies on 2D synthesized text fitting.}
    \colorbox{green1}{Forest}: the best performance.
    }
    \label{table:text_sample}
\end{table}

\begin{table}[t]
    \small
    \centering
    \begin{tabular}{l|cccc|c}
        \toprule
           & PSNR & SSIM & LPIPS &  MSE  & Time \\ 
        Strategies  & $\uparrow$ & $\uparrow$ & $\downarrow$ & $\downarrow$~(e-3) & $\downarrow$ (sec) \\ 
        \midrule
        PEMLP~\cite{pe} & 34.64 & 0.977 & 0.036 & 1.944 & 30.67 \\
        +EVOS & \cellcolor{green2}37.80 & \cellcolor{green2}0.983 &  \cellcolor{green2}0.025 & \cellcolor{green2}1.138  & 18.83\\ 
        \midrule
        SIREN~\cite{siren}  & 35.15 & 0.986 & 0.022 & 1.527 & 35.89 \\ 
        +EVOS  & \cellcolor{green2}37.42 & 0.985 & \cellcolor{green2}0.016 & \cellcolor{green2}1.002 & 24.15 \\
        \midrule
        GAUSS~\cite{gauss} & 36.97 & 0.986 & 0.013 & 1.946 & 50.44 \\
        +EVOS & \cellcolor{green2}39.28 & \cellcolor{green2}0.992 & \cellcolor{green2}0.005 & \cellcolor{green2}1.066 & 31.09\\ 
        \midrule
        FINER~\cite{finer} & 41.56 & 0.993 & 0.005 & 0.420 & 46.20 \\
        +EVOS & \cellcolor{green2}43.82 & 0.990 & 0.008 & \cellcolor{green2}0.349 & 28.01\\
        \bottomrule
    \end{tabular}
    \caption{
    {Quantitative comparison across different backbones on 2D text fitting task.}
    \colorbox{green2}{Mint}: enhance both in efficiency \& quality.
    }
    \label{table:text_backbone}
\end{table}

\section{2D Image Fitting Task}
\textbf{Settings.}
We evaluated EVOS on the widely used Kodak dataset~\cite{kodak}, which comprises 24 natural images at 768 $\times$ 512 resolution, distinct from the DIV2K dataset~\cite{div2k} used in Sec.~\ref{sec:exp}. All experimental settings, including hyperparameters, backbones, and network architectures, strictly followed those in Sec.~\ref{sec:compare_sota}. We compared reconstruction quality under fixed iterations to demonstrate our method's advantages.

\noindent \textbf{Results.}
As shown in Table~\ref{table:kodak}, our method achieves state-of-the-art efficiency compared to other sampling-based acceleration methods. Specifically, under fixed iterations, our approach reduces training time by 46.79\% while improving PSNR by 0.31 dB compared to standard training with constant scheduler, and achieves 29.79\% time reduction with 0.91 dB PSNR gains using step-wise scheduler. Notably, training with only 50\% of the data not only reduces computational cost but also improves per-iteration performance, further supporting our findings in Sec.~\ref{sec:compare_sota}.

\begin{table}[t]
    \small
    \centering
    \begin{tabular}{l|cc|cc|c}
        \toprule
          &  \multicolumn{2}{c|}{5k} & \multicolumn{2}{c|}{10k} & Time \\ 
          Settings & IoU~$\uparrow$ & CHD~$\downarrow$ & IoU~$\uparrow$ & CHD~$\downarrow$  & (sec) $\downarrow$ \\ 
        
        \midrule
        Standard &  0.949 & 1.45e-6 & \cellcolor{green1}0.967 & \cellcolor{green1}6.65e-7 & 171.26\\ 
        \midrule
        Uniform. & 0.918 & 3.66e-4 & 0.965 & 1.13e-3 & 96.07 \\
        \midrule
        INT~\cite{nmt} &  0.946 & 2.94e-6 & 0.962 & 1.96e-6 & 106.75\\ 
        \underline{INT}~\cite{nmt}\textsuperscript{$\star$} & 0.938 &1.65e-5 &  0.956 & 1.52e-5 & 147.09\\
        \underline{INT}~\cite{nmt})\textsuperscript{$\dagger$} &0.951 & 1.59e-6 & 0.965 &1.01e-6 & 115.28\\        
        \midrule 
        EVOS &  \cellcolor{green1}0.955 & \cellcolor{green1}1.42e-6 & 0.965 & 8.92e-7 & 98.31 \\ 
        \underline{EVOS} & \cellcolor{green1}0.955 & 1.44e-6 & \cellcolor{green1}0.967 & 8.18e-7 & 106.02 \\
        \bottomrule
        \multicolumn{6}{l}{\scriptsize{$\star$ denotes INT~(dense.)}} \\ 
        \multicolumn{6}{l}{\scriptsize{$\dagger$ denotes the best-performing variant reported in INT~\cite{nmt}.}} \\ 
    \end{tabular}
    \caption{
    {Comparison of sampling strategies on 3D shape fitting. CHD: CHamfer Distance. \colorbox{green1}{Forest}: the best performance.} 
    }
    \label{table:shape_sampler}
\end{table}

\begin{table*}[h]
    
    \centering
    \begin{tabular}{l|ccc|ccc|ccc|c}
        \toprule
                & \multicolumn{3}{c|}{1k Iterations} & \multicolumn{3}{c|}{2k Iterations} & \multicolumn{3}{c|}{5k Iterations} & \multicolumn{1}{c}{Time$\downarrow$} \\ 
       Strategies & PSNR $\uparrow$ &SSIM $\uparrow$ &LPIPS$\downarrow$ &  PSNR$\uparrow$ & SSIM$\uparrow$& LPIPS$\downarrow$ & PSNR$\uparrow$& SSIM$\uparrow$& LPIPS$\downarrow$& (sec) \\
        \midrule
        \rowcolor{gray2}
       Standard   & 30.54  &   0.848  &  0.235  & 33.47 &   0.906  &   0.131  & 36.10 &   0.938  &   0.074 & 266.55 \\ 
        \midrule
        Uniform.  &  29.91  &   0.832  &   0.266  &  32.55  &   0.891  &   0.161  &  35.16 &   0.927 &   0.096 & \cellcolor{green2}137.53 \\
        
        EGRA~\cite{egra}  &  29.92 &   0.831 &   0.267   & 32.63 &   0.891 &   0.159  &  35.21 &   0.927 &   0.095 & \cellcolor{green2}141.62\\ 
        
        Expan.~\cite{es}  & 30.47  &   0.844  &   0.237  &  33.015  &   0.897 &   0.139   &  35.30  &   0.927 &   0.084   & \cellcolor{green2}138.23 \\ 

        INT~\cite{nmt}~(dense.) & \cellcolor{green2}30.92  &  \cellcolor{green2}0.853  &   \cellcolor{green2}0.229   &  \cellcolor{green2}33.61  &   0.906  &   \cellcolor{green2}0.129   &  36.08  &   0.936  &  0.074 &230.76 \\
        
        INT~\cite{nmt}~(incre.) &  \cellcolor{green2}30.92 &   \cellcolor{green2}0.853 &  \cellcolor{green2}0.229   &  31.43  &  0.853 &  0.211  &  34.65  &   0.904 &   0.109 & \cellcolor{green2}198.88\\
        

        
       EVOS~(proposed)&  \cellcolor{green1}31.68  &  \cellcolor{green1}0.862  &   \cellcolor{green1}0.209   &  \cellcolor{green1}34.81  &   \cellcolor{green1}0.915 &   \cellcolor{green2}0.109   &  \cellcolor{green2}36.41  &   0.934  &  \cellcolor{green2}0.073 & \cellcolor{green2}141.85\\ 
        
        \midrule
        \underline{Uniform.}  & 29.28  &   0.814  &   0.301   &  32.25 &   0.885  & 0.175  &  35.941 &   0.935 &   0.080 & \cellcolor{green2}153.08 \\ 
        
       \underline{EGRA}~\cite{egra}  & 29.27  &  0.812  &   0.302  &  32.29 &   0.885  &   0.175  & 35.96 &  0.935 &   0.081 & \cellcolor{green2}166.54 \\
        
       \underline{Expan.}~\cite{es} & 30.46  &   0.839  &   0.241   & 33.11  &  0.895 &   0.148 & \cellcolor{green2}36.12 &   0.935  &   0.081   & \cellcolor{green2}153.02 \\ 

        \underline{INT}~\cite{nmt}~(dense) & \cellcolor{green2}31.07   &   0.845  &   0.250   &   \cellcolor{green2}33.58 &    0.901  &   0.140  &  36.00  &    0.935 &   0.078  & 251.08\\
        
        \underline{INT}~\cite{nmt}~(incre.)\textsuperscript{$\dagger$} & \cellcolor{green2}31.07  &   0.845 &   0.249   &  30.33  &  0.821  &  0.265   &  \cellcolor{green2}36.22  &   0.938 &   0.076 & \cellcolor{green2}177.05 \\


       \underline{EVOS~(proposed)} &   \cellcolor{green2}31.39  &   0.843 &   \cellcolor{green2}0.231  &  \cellcolor{green2}34.708  &   \cellcolor{green2}0.907  &   \cellcolor{green1}0.118   &  \cellcolor{green1}37.01 &  \cellcolor{green1}0.943  &   \cellcolor{green1}0.066  & \cellcolor{green2}187.15\\ 
        
        \bottomrule
        \multicolumn{8}{l}{\scriptsize{$\dagger$ denotes the best-performing variant reported in INT~\cite{nmt}.}}
    \end{tabular}
    
    \caption{
    Comparison of sampling strategies on Kodak datasets.
    Strategies without underlines employ constant scheduler ($\beta=0.5$), while
    \underline{underlined strategies} implement step-wise scheduler. 
    \colorbox{green1}{Forest}: the best performance; \colorbox{green2}{Mint}: exceeds standard training.
    }
    \label{table:kodak}
\end{table*}

\section{3D Shape Fitting Task}
\textbf{Backgound \& Settings.}
We used Signed Distance Fields~(SDF) to represent 3D shapes, a widely adopted approach in computer graphics~\cite{sdf-survey}. The fitting task can be formulated as $F_\theta(\mathbf{x}): (x,y,z) \mapsto (s)$, where $(x,y,z)$ represents the coordinate of given points and $s$ denotes the signed distance to the surface.
Following INT~\cite{nmt}, we employed an $8\times256$ MLP with SIREN architecture. We evaluated our method on the Asian Dragon scene from the Stanford 3D Scanning Repository~\cite{stanford-sdf-data}. The total iterations were set to $T=10,000$, with other settings remaining consistent with Sec.~\ref{sec:compare_sota}. Following~\cite{bacon}, we sampled points from the surface using coarse (Laplacian noise with variance 0.1) and fine (Laplacian noise with variance 0.001) sampling procedures, randomly selecting 50,000 points per iteration.
For EVOS, due to the absence of suitable high-frequency extractors for SDF, we temporarily disabled the crossover component. 
Given the varying degrees of task incompatibility exhibited by EGRA~\cite{egra}, Expan.~\cite{es}, and Soft Mining~\cite{soft}, we confined our experiment to standard training, uniform sampling, INT, and its variants.

\begin{figure}[t]
    \centerline{\includegraphics[width=\columnwidth]{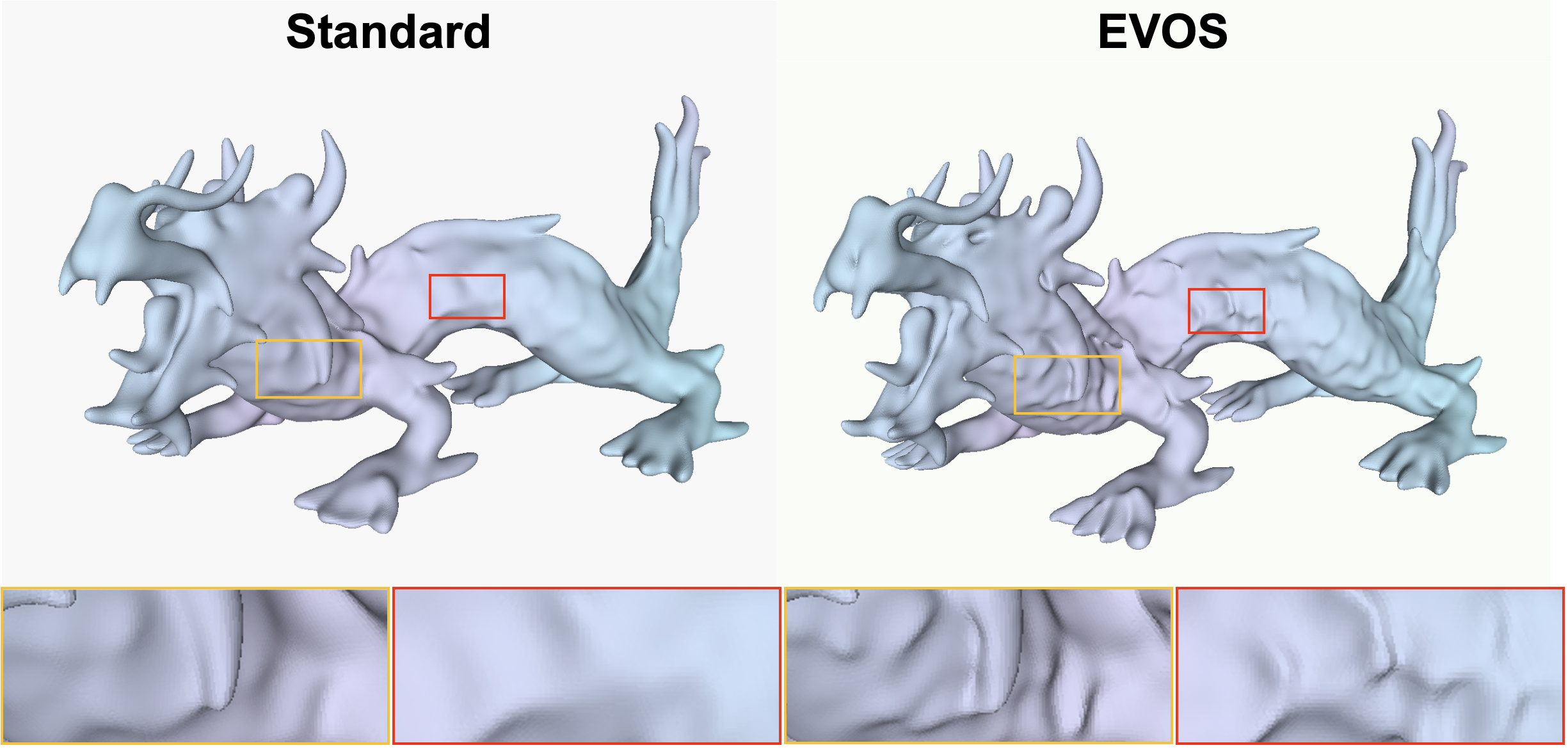}}
    \caption{Visual comparison of 3D shape fitting with fixed 90-second training.}
    \label{fig:shape_compare}
\end{figure}

\noindent \textbf{Results.} 
Table~\ref{table:shape_sampler} demonstrates that EVOS achieves significant efficiency improvements compared to standard training, INT, and its variants. Our method maintains comparable reconstructed quality while reducing training time by 38.10\%. At 5,000 iterations, we achieve improvement in IoU while maintaining acceleration benefits.
Visual comparisons in Fig.~\ref{fig:shape_compare} show reconstructed shapes after 90 seconds of training, demonstrating that EVOS significantly enhances reconstructed details.

\begin{table*}[t]
    \centering
    \begin{tabular}{l|ccc|ccc|ccc}
        \toprule
                & \multicolumn{3}{c|}{1k Iterations} & \multicolumn{3}{c|}{2k Iterations} & \multicolumn{3}{c}{5k Iterations} \\ 
        Settings & PSNR $\uparrow$ &SSIM $\uparrow$ &LPIPS$\downarrow$ &  PSNR$\uparrow$ & SSIM$\uparrow$& LPIPS$\downarrow$ & PSNR$\uparrow$& SSIM$\uparrow$& LPIPS$\downarrow$  \\
        \midrule
        \rowcolor{red2}
        Original &  25.98  &   0.815  &   0.141   &  25.98  &   0.824  &   0.095  &  25.97  &  0.830 &   0.071  \\
        \midrule
        \rowcolor{green2}
        w/o Fuzzy Indexing &  31.48  &   0.903  &   0.124   &  33.52  &   0.930  &   0.072   &  35.29  &   0.948  &   0.044  \\
        \midrule
        Hard~($\alpha=0$) & 30.42  &  0.877  &   0.159   & 33.07  &   0.920  &   0.071   &  35.25  &  0.944 &  0.046 \\ 
        $\alpha = 0.1$ &  30.67  &   0.884  &   0.148   &  33.21 &   0.923  &   0.069   &  35.27  &   0.946  &   0.045 \\ 
        $\alpha = 0.3$ & 31.15  &  0.896  &   0.131   &   33.45  &    0.928  &   0.067   &   35.27  &    0.948  &   0.048 \\ 
        \rowcolor{green2}
        $\alpha = 0.5$\textsuperscript{$\star$} & 31.49 & 0.903 & 0.124 & 33.52 & 0.931 & 0.072 & 35.31 & 0.948 & 0.045 \\ 
        $\alpha = 0.7$ & 31.47 &  0.903 &   0.125  &  33.44 &   0.930  &  0.074  &  35.18 & 0.947  &  0.047  \\ 
        $\alpha = 0.9$ & 31.17  &   0.899  &   0.131 &  33.18  &  0.928  &   0.079  & 34.79 &   0.944  &  0.052 \\
        Important~($\alpha = 1$) &  30.78  &  0.893  &   0.140   &  32.97  &   0.927  &   0.082   &  34.52  &   0.943  &    0.055 \\ 
        \midrule
        w/o warmup & 30.98  &  0.895  & 0.140  & 33.21 &  0.928 & 0.076  & 35.18  &   0.947  &  0.045 \\
        warmup for 0.1k & 31.34  &  0.901  &  0.131   &   33.32  &    0.928  &    0.076   &   35.15  &    0.946  &    0.047 \\
        warmup for 0.5k & 31.51  &   0.903  &  0.124   & 33.43  &  0.929  &   0.073   &  35.22  &   0.947  &   0.046 \\ 
        \bottomrule
        \multicolumn{10}{l}{\scriptsize{$\star$ The final $\alpha$ in our implementation.}} \\ 
    \end{tabular}
    \caption{{Results of empirical parameter tuning for {Soft Mining}. All experiments except \textit{Original} are conducted with the basic improvement of \textit{w/o Fuzzy Indexing}. Other experimental settings follow Sec.\ref{sec:compare_sota}.}}%
    \label{table:soft_tune}
\end{table*}

\section{Implementation of Soft Mining}
\textbf{Background.}
Soft Mining is an acceleration method for Neural Radiance Field~(NeRF) that utilizes Langevin Monte-Carlo sampling to form training batches during optimization. The sampling process can be formulated as:
\begin{equation}
\label{eq:lmc}
\mathbf{x}_{t+1}=\mathbf{x}_t+a \nabla \log Q\left(\mathbf{x}_t\right)+b \boldsymbol{\eta}_{t+1},
\end{equation}
where $\mathbf{x}_{t}$ represents sampled data at step $t$, $a$ and $b$ are hyperparameters, $Q(\mathbf{x})$ denotes the L1 norm of the error, and $\boldsymbol{\eta} \sim \mathcal{N}(\mathbf{0,1})$ is Gaussian noise.
To mitigate training bias introduced by importance sampling, they propose soft mining to regulate the loss function:
\begin{equation}
\mathcal{L}=\frac{1}{N} \sum_{n=1}^N\left[\frac{\operatorname{err}\left(\mathbf{x}_n\right)}{\operatorname{sg}\left(Q\left(\mathbf{x}_n\right)\right)^\alpha}\right], \quad \text { where } \alpha \in[0,1],
\end{equation}
where $\alpha$ controls mining softness ($\alpha=0$ for pure hard mining, $\alpha=1$ for pure importance mining), $\operatorname{err}(\mathbf{x})$ is the L2 norm of the error, $\operatorname{sg}(\cdot)$ is the stop gradient operator, and $N$ denotes the batch size. Further details are provided in the original paper~\cite{soft}.

\begin{table*}[h]
    \centering
    \begin{tabular}{l|ccc|ccc|ccc|c}
        \toprule
                & \multicolumn{3}{c|}{1k Iterations} & \multicolumn{3}{c|}{2k Iterations} & \multicolumn{3}{c|}{5k Iterations} & \multicolumn{1}{c}{Time$\downarrow$} \\ 
        Strategies & PSNR $\uparrow$ &SSIM $\uparrow$ &LPIPS$\downarrow$ &  PSNR$\uparrow$ & SSIM$\uparrow$& LPIPS$\downarrow$ & PSNR$\uparrow$& SSIM$\uparrow$& LPIPS$\downarrow$& (min)  \\
        \midrule
        Uniform. &  27.14  &   0.802 &   0.305  & 28.77 &  0.851  & 0.212  & 30.14 &  0.886  &  0.139 & 4.76 \\
        EGRA~\cite{egra} & 27.10  &   0.799  &   0.309   &   28.80  &    0.850  &   0.214  &   30.32  &   0.885  &   0.140  & 6.17\\ 
        Expan.~\cite{es} & 27.21  &   0.790  &  0.303  &  28.47  &  0.831  &   0.241   &  29.67  &  0.862  &  0.183 & 5.05 \\
        Soft Mining~\cite{soft} &  \cellcolor{green1}27.80  &   \cellcolor{green1}0.819  &   \cellcolor{green1}0.274   &  29.29  &   \cellcolor{green1}0.859  &    \cellcolor{green1}0.137 & \cellcolor{green1}30.73  &  \cellcolor{green1}0.888  &  \cellcolor{green1}0.136 & 6.94 \\ 
        EVOS & 26.22  &   0.742  &   0.308   &  \cellcolor{green1}29.49  &   0.847  &   0.166   & 29.74  &  0.847  &   0.154  & 6.87\\
        
        \bottomrule
    \end{tabular}
    \caption{
    Comparison of sampling strategies under extremely ultra-low selection ratio ($\beta=0.05$). \colorbox{green1}{Green}: the best performance. 
    }
    \label{table:compare_0.05}
\end{table*}

\begin{table*}[h]
    \centering
    \begin{tabular}{l|ccc|ccc|ccc}
        \toprule
                & \multicolumn{3}{c|}{5 minutes} & \multicolumn{3}{c|}{15 minutes } & \multicolumn{3}{c}{25 minutes}  \\ 
        Strategies & PSNR $\uparrow$ &SSIM $\uparrow$ &LPIPS$\downarrow$ &  PSNR$\uparrow$ & SSIM$\uparrow$& LPIPS$\downarrow$ & PSNR$\uparrow$& SSIM$\uparrow$& LPIPS$\downarrow$ \\
        \midrule

        Soft.~\cite{soft}~($\beta=0.05$) & 30.19 &  0.878 &  0.156 & 31.57 &  0.901 & 0.106 &  32.01 & 0.907 &  0.092 \\
        
        Soft.~\cite{soft}~($\beta=0.5$)& 31.48 & 0.903 &  0.124  & 34.02 & 0.936 & 0.062 & 35.14 &  0.947 & 0.046 \\
        \rowcolor{green1}
        \cellcolor{white}EVOS~($\beta=0.5$) & 32.69 &  0.912 & 0.082 & 36.43 & 0.954 &  0.026 & 37.50 & 0.961 & 0.019 \\ 
        \bottomrule
    \end{tabular}
    \caption{
    {{Comparison of sampling strategies under fixed time budget.} Reported times represent cumulative training duration across all dataset samples. All experimental settings follow Sec.~\ref{sec:compare_sota}.
    \colorbox{green1}{Green}: the best performance. 
    }%
    }
    \label{table:evos_soft_comp}
\end{table*}

\noindent \textbf{Parameter Tuning \& Ablation Study.}
The official implementation of Soft Mining showed significant performance degradation when applied to natural image fitting (Sec.~\ref{sec:compare_sota}), likely due to fundamental differences between image fitting and radiance field synthesis tasks. 
To ensure fair comparison, we conducted comprehensive parameter tuning and ablation studies.
Our investigation focused on three key components: the softness parameter $\alpha$, warmup iteration count, and fuzzy indexing mechanism. Parameters $a$ and $b$ exhibited minimal influence on performance; hence, we retained their default values. Results in Table~\ref{table:soft_tune} demonstrate that fuzzy indexing significantly degraded performance. Based on empirical analysis, we determined optimal settings by disabling fuzzy indexing, setting $\alpha=0.5$, and maintaining the original 1,000 warmup iterations.

\noindent \textbf{Fuzzy Indexing Issue.}
Fuzzy Indexing serves as an engineering preprocessing step rather than an algorithmic component of soft mining.
After LMC (Eq.~\ref{eq:lmc}), sampled points~(coordinates) are processed into bounded values $w\in[0,1]$ and subsequently scaled to the INR coordinate space, typically through min-max normalization to [-1,1].
This process can result in sampling coordinates beyond the scope of available ground truth values. For instance, sampled coordinates (5.52, 9.27) lack corresponding ground truth, which is only available at discrete points like (5,9) or (6,10), potentially lowering supervision accuracy. This fuzzy indexing issue is particularly pronounced with limited training data, leading to significant performance degradation in image fitting tasks while maintaining effectiveness in NeRF applications.

\noindent \textbf{How to Disable Fuzzy Indexing?}
To mitigate the performance degradation caused by fuzzy indexing issues, we disabled this step by regulating $w$ to real coordinate value.
Specifically, after obtaining sampled value $w$, we first transform it to coordinate space and round it to the nearest integer (corresponding to real coordinates) before applying INR's min-max normalization. 
As shown in Table~\ref{table:soft_tune}, reconstruction quality improves significantly after disabling fuzzy indexing. More details can be found in our code.

\noindent \textbf{Analysis for Degraded Performance.}
As shown in Table~\ref{table:compare_0.5}, despite our optimized implementation, Soft Mining's performance remains unsatisfactory, particularly in later training stages, falling below uniform sampling. We attribute this limitation to LMC sampling mechanism's incompatibility with fitting tasks involving smaller sampling sets.
A key distinction between general INR training~(e.g., audio, text, image and shape fitting task) and NeRF training lies in their training data volume. NeRF training typically requires $N \sim 10^{10}$ points, calculated as:
\begin{equation}
    N = \underbrace{(128+64)}_{\text{points along a ray}}\times \underbrace{1080 \times 768}_{\text{pixels}} \times \underbrace{100}_{\text{views}} \sim 10^{10}.
\end{equation}
In standard NeRF training, each iteration uniformly samples $4096\times(128+64) \sim 10^5$ points per batch, representing 0.005\% of total training data.
Conversely, a 1080P image contains only $N \sim 10^5$ points, allowing INR fitting to access all training data in each iteration without batch splitting. This smaller sampling space and larger sampling ratio conflict with LMC sampling's design paradigm, potentially explaining Soft Mining's degraded performance in our experiments.
To validate this hypothesis, we set sampling ratio $\beta=0.05$ to simulate batch training in image fitting. Results in Table~\ref{table:compare_0.05} support our hypothesis, with Soft Mining outperforming other methods, including EVOS, under ultra-low sampling ratios. However, as shown in Table~\ref{table:evos_soft_comp}, despite this advantage, Soft Mining's performance remains inferior to EVOS under equivalent time budgets for general INR acceleration.

\section{Implementation of EGRA \& Expan.}
We reimplemented EGRA~\cite{egra} and Expansive Supervision~(Expan.)~\cite{es} following their algorithmic designs and hyperparameter settings, as official implementations were unavailable. More details can be found in our released code.

\section{Compatibility for Different Scheduler}
\textbf{Settings.}
We further evaluated EVOS with linear and cosine increment schedulers. The linear scheduler increases the selection ratio from 0\% to 100\% across iterations, with selection intensity $q = \frac{N}{T}$, where $T$ denotes total iterations and $N$ represents total coordinates. The cosine scheduler implementation follows~\cite{nmt}. All other experimental settings remain consistent with Sec.~\ref{sec:compare_sota}.

\noindent \textbf{Results.}
Results for linear and cosine schedulers are presented in Table~\ref{table:linear_strategy} and Table~\ref{table:cosine_strategy}, respectively. With the linear increment scheduler, EVOS reduces training time by 36.24\% while achieving a 1.26 dB gain in PSNR. Similarly, with the cosine increment scheduler, it achieves a 36.49\% reduction in training time with a 1.20 dB PSNR improvement. These results demonstrate EVOS's robust performance across different scheduling strategies.

\begin{table*}[h]
    
    \centering
    \begin{tabular}{l|ccc|ccc|ccc|c}
        \toprule
                & \multicolumn{3}{c|}{1k Iterations} & \multicolumn{3}{c|}{2k Iterations} & \multicolumn{3}{c|}{5k Iterations} & \multicolumn{1}{c}{Time$\downarrow$} \\ 
        Strategies & PSNR $\uparrow$ &SSIM $\uparrow$ &LPIPS$\downarrow$ &  PSNR$\uparrow$ & SSIM$\uparrow$& LPIPS$\downarrow$ & PSNR$\uparrow$& SSIM$\uparrow$& LPIPS$\downarrow$& (min)  \\
        \midrule
        \rowcolor{gray2}
        Standard   & 31.06 & \cellcolor{green1}0.899 & 0.123 & 34.34 & 0.944 & 0.042 & 37.10 & 0.964 & 0.021 & 180.45 \\ 
        \midrule
        Uniform.  & 29.84  &   0.874  &   0.175  &  33.23 &  0.931  &   0.062   & 37.20  &  0.963  &   0.021 & 111.30\\
        
        EGRA~\cite{egra} &   29.85  &   0.873 &   0.175  &  33.27  &   0.930  &   0.062  &  37.11  &   0.963  &   0.021   & 113.21 \\ 
        Expan.~\cite{es} & 30.80  &    0.888  &   0.133   &  33.79  &   0.934  &   0.053   &  37.19  &   0.962  &   0.023   & 136.23 \\ 
        
        \midrule
        INT~\cite{nmt}~(incre.) & 31.68  &   0.889  &  0.129   &  31.79  &   0.890  &   0.103   &  37.23  &   0.963  &  0.021  & 129.41\\

        INT~\cite{nmt}~(dense.)  & 31.68  &   0.889  &   0.129   &  34.55  &   0.937  &   0.052   &  37.21  &   0.963  &   \cellcolor{green1}0.019   & 177.38 \\
        \midrule
        
        EVOS~(proposed)&  \cellcolor{green1}32.26  &   0.894  &   \cellcolor{green1}0.094   &  \cellcolor{green1}35.95  &   \cellcolor{green1}0.945  &   \cellcolor{green1}0.038   &  \cellcolor{green1}38.36  &   \cellcolor{green1}0.968  &  \cellcolor{green1}0.019   & 115.05\\ 
        
        \bottomrule
    \end{tabular}
    \caption{
    {Comparison of sampling strategies with \textbf{linear} increment scheduler.} \colorbox{green1}{Green}: the best performance.
    }
    \label{table:linear_strategy}
\end{table*}

\begin{table*}[h]
    \centering
    \begin{tabular}{l|ccc|ccc|ccc|c}
        \toprule
                & \multicolumn{3}{c|}{1k Iterations} & \multicolumn{3}{c|}{2k Iterations} & \multicolumn{3}{c|}{5k Iterations} & \multicolumn{1}{c}{Time$\downarrow$} \\ 
        Strategies & PSNR $\uparrow$ &SSIM $\uparrow$ &LPIPS$\downarrow$ &  PSNR$\uparrow$ & SSIM$\uparrow$& LPIPS$\downarrow$ & PSNR$\uparrow$& SSIM$\uparrow$& LPIPS$\downarrow$& (min)  \\
        \rowcolor{gray2}
        \midrule
        Standard   & 31.06 & \cellcolor{green1}0.899 & 0.123 & 34.34 & \cellcolor{green1}0.944 & 0.042 & 37.10 & 0.964 & 0.021 & 180.45 \\
        \midrule
        Uniform.  &  29.54  &   0.868  &   0.188  &  33.00  &   0.928  &   0.068   &  37.11  &   0.962  &  0.023 & 110.81\\
        
        EGRA~\cite{egra} & 29.54  &   0.866  &  0.189  &  33.08  &  0.928  &  0.066  &  37.27  &   0.964 &  0.0194 & 112.88 \\

        Expan.~\cite{es} & 30.53  &   0.881  &   0.140  &  33.63 &   0.931 &   0.056   & 37.34  &   0.963  &   0.020 &126.98 \\ 
        
        \midrule
        INT~\cite{nmt}~(incre.) & \cellcolor{green1}31.68  &    0.882  &    0.132   &   30.81  &   0.869 &    0.138  &   37.22 &    0.963  &  0.019  & 128.77\\

        INT~\cite{nmt}~(dense.)  &  \cellcolor{green1}31.68  &   0.882  &   0.132   &  34.67  &  0.935  &  0.055   &  37.17 &   0.962  &   0.020 & 175.95 \\
        \midrule

        EVOS~(proposed) &   31.57 &   0.881 &   \cellcolor{green1}0.106 & \cellcolor{green1} 35.63 &   0.941 &   \cellcolor{green1}0.041 &  \cellcolor{green1}38.30 &   \cellcolor{green1}0.968  &  \cellcolor{green1}0.016 & 114.60 \\ 
        \bottomrule
    \end{tabular}
    \caption{
    {Comparison of sampling strategies with \textbf{cosine} increment scheduler.} \colorbox{green1}{Green}: the best performance. 
    }
    \label{table:cosine_strategy}
\end{table*}

\section{Compatibility for Different Selection Ratio}
\textbf{Settings.} 
We investigated the impact of selection ratio $\beta$ under constant scheduler by evaluating $\beta = \{0.3, 0.7\}$. Results for $\beta = 0.5$ are presented in Table~\ref{table:compare_0.5}~(without underline). All other experimental settings remain consistent with Sec.~\ref{sec:compare_sota}.

\noindent \textbf{Results.}
Results for selection ratios of 30\% and 70\% are presented in Table~\ref{table:compare_0.3} and Table~\ref{table:compare_0.7}, respectively. Our method maintains efficiency improvements with 70\% selection ratio; however, performance degrades with a lower sampling intensity (30\%), where despite greater time reduction, it fails to surpass standard training performance.

\begin{table*}[h]
    \centering
    \begin{tabular}{l|ccc|ccc|ccc|c}
        \toprule
                & \multicolumn{3}{c|}{1k Iterations} & \multicolumn{3}{c|}{2k Iterations} & \multicolumn{3}{c|}{5k Iterations} & \multicolumn{1}{c}{Time$\downarrow$} \\ 
        Strategies & PSNR $\uparrow$ &SSIM $\uparrow$ &LPIPS$\downarrow$ &  PSNR$\uparrow$ & SSIM$\uparrow$& LPIPS$\downarrow$ & PSNR$\uparrow$& SSIM$\uparrow$& LPIPS$\downarrow$& (min)  \\
        \midrule
        \rowcolor{gray2}
        Standard   & 31.06 & \cellcolor{green1}0.899 & 0.123 & 34.34 & \cellcolor{green1}0.944 & 0.042 & \cellcolor{green1}37.10 & \cellcolor{green1}0.964 & \cellcolor{green1}0.021 & 180.45 \\
        
        \midrule
        Uniform. &  29.71  &  0.872  &   0.175   &  32.42  &   0.921  &  0.078  &  34.92  &  0.946  &   0.037 &58.24\\
        
        EGRA~\cite{egra} & 29.73  &  0.871  &   0.175   &  32.49  &   0.920  &   0.077   &  35.01 &   0.946  &   0.037  & 61.24 \\ 
         
        Expan.~\cite{es} & 30.60  &   0.883  &   0.125   & 32.92  &   0.920  &  0.061   &  35.09  &   0.943  &   0.035 & 58.46 \\ 

        Soft.~\cite{soft}  & 30.77 &  0.891  &   0.146   &  32.68  &  0.921  &   0.087   &  34.43  &   0.941 &   0.055  & 66.31  \\
        
        \midrule
        INT~\cite{nmt}~(incre.) & 31.74  &   0.890  &   0.121   &  26.62  &   0.738  &   0.289   &  29.45  &  0.832  &   0.162  & 85.13 \\

        INT~\cite{nmt}~(dense.)  & 31.74  &   0.890  &   0.121   &  34.40  &  0.926  &   0.062   & 36.72  &   0.951  &  0.037 & 125.17 \\
        \midrule
        
        
        EVOS~(proposed) & \cellcolor{green1}31.87 &   0.891  &   \cellcolor{green1}0.097   &  \cellcolor{green1}35.14  &   0.939  &   \cellcolor{green1}0.037   &  36.24  &   0.948  &  0.029 & 63.71\\ 
        
        \bottomrule
    \end{tabular}
    \caption{
    Comparison of sampling strategies with \textbf{constant} increment scheduler~{($\beta=0.3$)}. \colorbox{green1}{Green}: the best performance.
    }
    \label{table:compare_0.3}
\end{table*}

\begin{table*}[h]
    
    \centering
    \begin{tabular}{l|ccc|ccc|ccc|c}
        \toprule
                & \multicolumn{3}{c|}{1k Iterations} & \multicolumn{3}{c|}{2k Iterations} & \multicolumn{3}{c|}{5k Iterations} & \multicolumn{1}{c}{Time$\downarrow$} \\ 
        Strategies & PSNR $\uparrow$ &SSIM $\uparrow$ &LPIPS$\downarrow$ &  PSNR$\uparrow$ & SSIM$\uparrow$& LPIPS$\downarrow$ & PSNR$\uparrow$& SSIM$\uparrow$& LPIPS$\downarrow$& (min)  \\
        \midrule
        \rowcolor{gray2}
        Standard   & 31.06 & 0.899 & 0.123 & 34.34 & 0.944 & 0.042 & 37.10 & 0.964 & 0.021 & 180.45 \\ 
        \midrule
        Uniform  & 30.72   &  0.893  &   0.134   &  33.89  &   0.939  &   0.048   &  36.85  &   0.961 &   0.021 & 126.83\\ 
        
        EGRA~\cite{egra} & 30.75  &   0.892  &   0.134 &  33.96 &   0.939  &   0.047  &  36.92 &   0.961  &  0.020 & 129.30   \\ 

        Expan.~\cite{es} & 31.28  &   0.899  &   0.116   &  34.34  &   0.940  &  0.042   &  37.10  &   0.961  & \cellcolor{green1}0.019 & 127.76  \\ 
        Soft.~\cite{soft}  & 31.89  &   0.908 &   0.113  &  33.98  &   0.935 &   0.065   & 35.79  &   0.952   &   0.040  & 139.05 \\
          
        \midrule
        INT~\cite{nmt}~(incre.)  & 31.37  &   0.902 &   0.118   &  33.81  &   0.929  &   0.057   &  37.17  &   0.957  &   0.027 & 143.86 \\

        INT~\cite{nmt}~(dense.)  & 31.37  &   0.902  &  0.118   &  34.42  &   0.941  &   0.0477   &  36.93  &  0.956  &  0.029  & 233.10 \\
        \midrule
        
        
        EVOS~(proposed) &  \cellcolor{green1}32.87  &   \cellcolor{green1}0.919  &   \cellcolor{green1}0.078   &  \cellcolor{green1}36.27  &   \cellcolor{green1}0.956  &   \cellcolor{green1}0.027  &  \cellcolor{green1}37.96  &   \cellcolor{green1}0.965  &   \cellcolor{green1}0.019 & 133.05\\ 
        
        \bottomrule
    \end{tabular}
    \caption{
    Comparison of sampling strategies with \textbf{constant} increment scheduler~{($\beta=0.7$)}. \colorbox{green1}{Green}: the best performance.
    }
    \label{table:compare_0.7}
\end{table*}

\section{Empirical Study for Designing $\Gamma(t)$}
In Section~\ref{fitness}, we introduced a linear increasing scheduler (Eq.~\ref{eq:eval_inter}) for key iterations to balance performance and efficiency, motivated by the observed gradual linear increase in distribution changes of $\mathcal{D}(F_\theta(\mathbf{x}), \mathbf{y})$ across iterations.
Here, we detail our analysis of the fitness distribution dynamics that guided the design of $\Gamma(t)$.

\begin{figure}[t]
    \centerline{\includegraphics[width=\columnwidth]{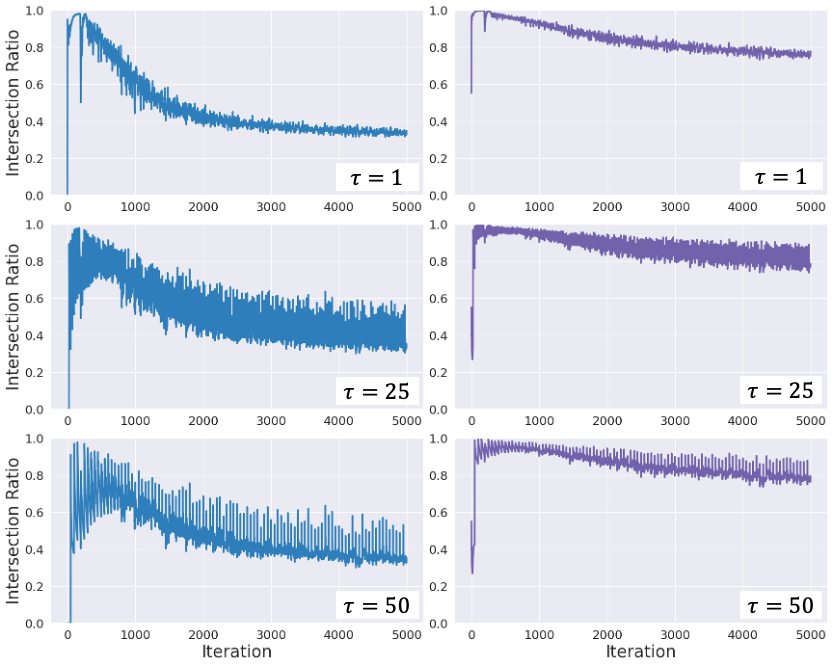}}
    \caption{$\mathcal{G}_t(\tau, \sigma)$ curves across iterations with $\tau=\{1,25,50\}$ and $\sigma=\{0.1, 0.5\}$.
    \textbf{\textcolor{audio_blue}{Blue}} and \textbf{\textcolor{audio_purple}{Purple}} curves represent $\sigma=0.1$ and $\sigma=0.5$, respectively.
    }
    \label{fig:study_interval}
\end{figure}

\noindent \textbf{Measuring Distribution Changes in $\mathcal{D}(F_\theta(\mathbf{x}), \mathbf{y})$.}
We define the distribution change of $\mathcal{D}_t(F_\theta(\mathbf{x}), \mathbf{y})$ at step $t$ as $\mathcal{G}_t(\tau, \sigma)$:
\begin{equation}
    \mathcal{G}_t(\tau, \sigma) = \frac{| \tilde{\mathbf{x}}^\sigma_t  \cup \tilde{\mathbf{x}}^\sigma_{t-\tau} |}{|\tilde{\mathbf{x}}^\sigma_t|},
\end{equation}
where
\begin{equation}
\tilde{\mathbf{x}}^\sigma_t = \underset{\{\tilde{\mathbf{x}}\}_{\sigma N} \subseteq \{\mathbf{x}\}_N} {\arg\max} ( \underbrace{||F_\theta(\mathbf{x})-\mathbf{y}||}_{\mathcal{D}_t(F_\theta(\mathbf{x}), \mathbf{y})}).
\end{equation}
Here, $\tau$ represents the measurement interval, $\sigma \in (0,1)$ denotes the sampling intensity for measurement, and $N$ is the total number of coordinates in set $\mathbf{x}$. 
A higher $\mathcal{G}_t(\tau, \sigma)$ indicates a larger intersection ratio, implying minimal distribution change, and vice versa.

\noindent \textbf{Settings \& Results.}
We conducted experiments using standard training without sampling-based acceleration, following the settings in Sec.~\ref{sec:compare_sota}. We plotted $\mathcal{G}_t(\tau, \sigma)$ curves across iterations with $\tau=\{1,25,50\}$ and $\sigma=\{0.1, 0.5\}$. 
Results in Fig.~\ref{fig:study_interval} demonstrate that $\mathcal{G}_t(\tau, \sigma)$ exhibits a gradual linear decrease across various settings as iterations progress. This increasing trend in distribution changes of $\mathcal{D}(F_\theta(\mathbf{x}), \mathbf{y})$ validates our design choice for $\Gamma(t)$ in Eq.~\ref{eq:eval_inter}.


\end{document}